\RequirePackage{fix-cm}
\documentclass[10pt,twocolumn,letterpaper]{article}

\usepackage{arxiv}
\usepackage{multirow}
\usepackage{algorithm}
\usepackage{algpseudocode}
\usepackage{colortbl}
\usepackage{tabularx}
\usepackage{array}
\usepackage{tikz}
\usepackage{microtype}
\usetikzlibrary{arrows.meta,calc,fit,backgrounds,positioning,shapes.geometric}

\definecolor{linkblue}{rgb}{0.15,0.40,0.68}
\usepackage[breaklinks,colorlinks,allcolors=linkblue]{hyperref}

\definecolor{qlbblue}{RGB}{0,114,178}
\definecolor{qlborange}{RGB}{230,159,0}
\definecolor{qlbsky}{RGB}{86,180,233}
\definecolor{qlbverm}{RGB}{213,94,0}
\definecolor{qlbgreen}{RGB}{0,158,115}
\definecolor{ours}{HTML}{EAF1FB}
\definecolor{fam}{HTML}{F2F2F2}
\definecolor{hvInk}{HTML}{182D45}
\definecolor{hvMuted}{HTML}{465C73}
\definecolor{hvRule}{HTML}{D5DFE8}
\definecolor{hvPaper}{HTML}{F5F7FA}
\definecolor{hvBlue}{HTML}{075C86}
\definecolor{hvSky}{HTML}{EAF4FA}
\definecolor{hvTeal}{HTML}{086B5D}
\definecolor{hvMint}{HTML}{E5F3EE}
\definecolor{hvAmber}{HTML}{994E2A}
\definecolor{hvSand}{HTML}{FBEEE1}
\definecolor{hvPlum}{HTML}{6B4BA8}
\definecolor{hvLilac}{HTML}{EFEAFA}
\definecolor{hvGrey}{HTML}{A9B6C4}

\tikzset{
  hv canvas/.style={
    x=1pt, y=1pt, inner sep=0pt, outer sep=0pt, text=hvInk,
    line cap=round, line join=round, >={Stealth[length=4pt,width=3.8pt]}},
  hv title/.style={font=\sffamily\bfseries\fontsize{10.4}{11.8}\selectfont},
  hv label/.style={font=\sffamily\fontsize{8.2}{9.6}\selectfont,text=hvMuted},
  hv bold/.style={hv label,font=\sffamily\bfseries\fontsize{8.2}{9.6}\selectfont},
  hv h1/.style={font=\sffamily\bfseries\fontsize{9.6}{11}\selectfont},
  hv h2/.style={font=\sffamily\bfseries\fontsize{8.4}{9.8}\selectfont},
  hv body/.style={font=\sffamily\fontsize{7.8}{9.2}\selectfont,text=hvMuted},
  hv tiny/.style={font=\sffamily\fontsize{6.6}{7.8}\selectfont,text=hvMuted},
  hv tinyb/.style={font=\sffamily\bfseries\fontsize{6.6}{7.8}\selectfont},
  hv kicker/.style={font=\sffamily\bfseries\fontsize{6.8}{8}\selectfont},
  hv panel/.style={fill=hvPaper,rounded corners=4pt},
  hv panel blue/.style={fill=hvSky,rounded corners=4pt},
  hv panel mint/.style={fill=hvMint,rounded corners=4pt},
  hv panel sand/.style={fill=hvSand,rounded corners=4pt},
  hv card/.style={fill=white,draw=hvRule,line width=.5pt,rounded corners=3.4pt},
  hv chip/.style={hv tinyb,text=#1,draw=#1!45,fill=#1!8,line width=.5pt,
    rounded corners=2.6pt,inner xsep=3.2pt,inner ysep=2.1pt},
  hv chip/.default=hvMuted,
  hv solidchip/.style={hv tinyb,text=white,fill=#1,rounded corners=2.6pt,
    inner xsep=3.4pt,inner ysep=2.2pt},
  hv solidchip/.default=hvTeal,
  hv flow/.style={-{Stealth[length=3.4pt,width=3.2pt]},line width=.8pt,draw=#1},
  hv flow/.default=hvMuted,
  hv icon/.style={draw=#1,line width=.9pt},
  pics/hv chart/.style={code={
    \draw[hv icon=hvBlue,rounded corners=1.4pt] (-5,-4.5) rectangle (5,4.5);
    \draw[hv icon=hvBlue] (-2.8,-2.2) -- (-.7,.4) -- (1.1,-.4) -- (3.2,2.1);
  }},
  pics/hv filter/.style={code={
    \draw[hv icon=hvBlue] (-4.6,3.8) -- (4.6,3.8) -- (1.4,-.4)
      -- (1.4,-3.3) -- (-1.4,-4.4) -- (-1.4,-.4) -- cycle;
  }},
  pics/hv book/.style={code={
    \draw[hv icon=hvBlue] (0,2.7) .. controls (-1.5,4) and (-3.9,4.3) .. (-5.3,3.6)
      -- (-5.3,-3.5) .. controls (-3.4,-2.7) and (-1.2,-2.8) .. (0,-4)
      .. controls (1.2,-2.8) and (3.4,-2.7) .. (5.3,-3.5) -- (5.3,3.6)
      .. controls (3.9,4.3) and (1.5,4) .. (0,2.7) -- (0,-4);
  }},
  pics/hv clock/.style={code={
    \fill[hvAmber!10] (0,0) circle (6.3);
    \draw[hv icon=hvAmber] (0,-.3) circle (3.8) (0,2) -- (0,-.3) -- (1.7,-1.3)
      (-1,5) -- (1,5);
  }},
  pics/hv clock/.default=hvAmber,
  pics/hv loop/.style={code={
    \draw[hv icon=hvAmber,-{Stealth[length=2.7pt,width=2.5pt]}]
      (25:3.5) arc[start angle=25,end angle=335,radius=3.5];
  }},
  pics/hv cross/.style={code={
    \draw[hv icon=hvAmber] (-2,-2) -- (2,2) (-2,2) -- (2,-2);
  }},
  pics/hv cross/.default=hvAmber,
  pics/hv check/.style={code={
    \draw[hv icon=hvTeal,line width=1.1pt] (-2.7,-.2) -- (-.8,-2.2) -- (3,2.4);
  }},
  pics/hv check/.default=hvTeal,
  pics/hv scissors/.style={code={
    \draw[hv icon=hvTeal] (-2.6,2.6) circle (1.6) (-2.6,-2.6) circle (1.6)
      (-1.2,1.8) -- (4.3,-2.9) (-1.2,-1.8) -- (4.3,2.9);
  }},
  pics/hv scissors/.default=hvTeal,
  pics/hv probe/.style={code={
    \draw[draw=#1,line width=.95pt] (0,.8) circle (3.5);
    \draw[draw=#1,line width=1.15pt] (2.5,-1.7) -- (5,-4.2);}},
  pics/hv probe/.default=hvPlum,
  pics/hv gold/.style={code={
    \draw[draw=#1,line width=.85pt] (0,0) circle (4.4) (0,0) circle (1.9);
    \fill[#1] (0,0) circle (.85);}},
  pics/hv gold/.default=hvTeal,
  pics/hv archive/.style={code={
    \draw[hv icon=hvBlue] (0,3.9) ellipse (5.5 and 1.9)
      (-5.5,3.9) -- (-5.5,-3.7) arc[start angle=180,end angle=360,x radius=5.5,y radius=1.9]
      -- (5.5,3.9)
      (-5.5,.1) arc[start angle=180,end angle=360,x radius=5.5,y radius=1.9];
  }},
  pics/hv chip/.style={code={
    \draw[hv icon=hvBlue,rounded corners=1pt] (-4.8,-4.8) rectangle (4.8,4.8)
      (-2,-2) rectangle (2,2);
    \foreach \p in {-2.8,0,2.8}{
      \draw[hv icon=hvBlue] (\p,4.8) -- (\p,6.6) (\p,-4.8) -- (\p,-6.6)
        (4.8,\p) -- (6.6,\p) (-4.8,\p) -- (-6.6,\p);
    }
  }},
  pics/hv lock/.style={code={
    \draw[hv icon=hvBlue,rounded corners=1pt] (-4.4,-4.8) rectangle (4.4,1.3);
    \draw[hv icon=hvBlue] (-2.5,1.4) -- (-2.5,3.7)
      arc[start angle=180,end angle=0,radius=2.5] -- (2.5,1.4)
      (0,-1) -- (0,-2.4);
  }},
  pics/hv student/.style={code={
    \draw[hv icon=hvTeal,line width=1.1pt] (-9,2.5) -- (0,6.3) -- (9,2.5)
      -- (0,-1.3) -- cycle (5.5,1) -- (5.5,-3.4)
      .. controls (3.6,-5.8) and (-3.6,-5.8) .. (-5.5,-3.4) -- (-5.5,1)
      (9,2.5) -- (9,-2.5);
  }},
  pics/hv prefix/.style={code={
    \fill[hvBlue!13,rounded corners=2pt] (0,-5.5) rectangle (49,5.5);
    \foreach \x in {3,14.7,26.4,38.1}{
      \fill[hvBlue!76,rounded corners=1pt] (\x,-2.55) rectangle ++(8,5.1);
    }
  }},
  pics/hv tail/.style={code={
    \fill[hvRule!44,rounded corners=2pt] (0,-5.5) rectangle (56,5.5);
    \foreach \x in {3,13.7,24.4,35.1,45.8}{
      \fill[hvRule!86!hvMuted,rounded corners=1pt] (\x,-2.55) rectangle ++(7,5.1);
    }
  }},
  pics/hv cue/.style={code={
    \path[fill=white,draw=hvBlue,line width=.65pt,rounded corners=1.8pt]
      (0,7.2) -- (54,7.2) -- (54,-7.2) -- (12,-7.2)
      -- (5,-10.3) -- (6,-7.2) -- (0,-7.2) -- cycle;
    \node[hv bold,text=hvBlue] at (27,0) {Answer now};
  }},
  pics/hv ending/.style={code={
    \path[fill=hvMint,draw=hvTeal!35,line width=.45pt,rounded corners=2pt]
      (0,-5.5) rectangle (31,5.5) (34,-5.5) rectangle (88,5.5);
    \draw[hv icon=hvTeal,line width=.7pt] (4.6,-1.3) rectangle (7.3,1.4);
    \node[hv bold,text=hvTeal] at (19.5,0) {Stop};
    \node[hv bold,text=hvTeal] at (61,0) {Gold answer};
  }},
}

\newcommand{\method}{PRFT}
\newcommand{\methodlong}{Prefix Rejection Fine-Tuning}

\arxivheader{%
  \includegraphics[height=8.2mm]{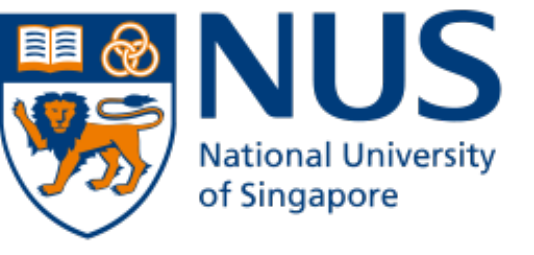}%
  \hfill%
  \includegraphics[height=8.5mm]{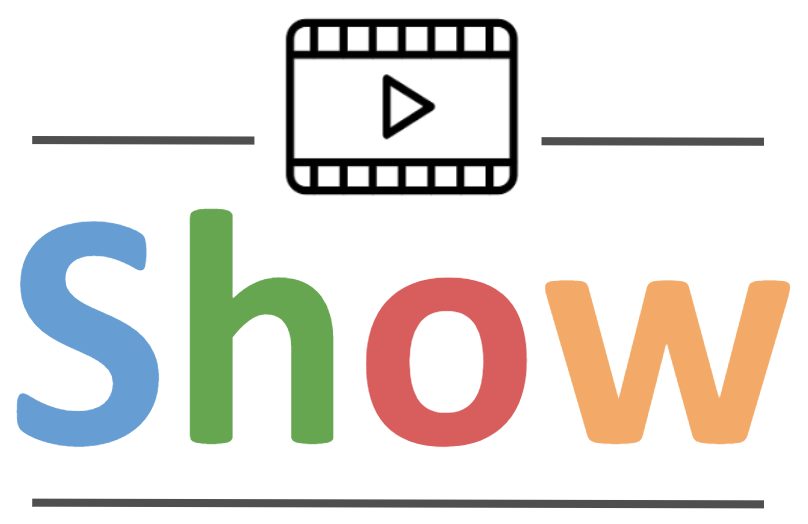}%
}

\title{\method{}: \methodlong{} for\\Multimodal Reasoning Distillation}

\author{%
  Binjie Zhang$^{1,*}$\quad
  Jiahao Tang$^{2,*}$\quad
  Alex Jinpeng Wang$^{2,\dagger}$\quad
  Mike Zheng Shou$^{1,\dagger}$%
}

\affiliation{%
  $^{1}$Show Lab, National University of Singapore \qquad
  $^{2}$Central South University%
}

\authornote{%
  $^{*}$Equal contribution \qquad $^{\dagger}$Corresponding authors%
}

\begin{document}
\maketitle
\begin{abstract}
In multimodal reasoning distillation, post-training methods such as supervised fine-tuning (SFT) enable a student vision-language model (VLM) to learn stronger reasoning capabilities from a teacher model.
However, on complex figures, teacher trajectories decouple visual deduction from text termination: the teacher resolves key visual cues early in the reasoning prefix, whereas fine-grained visual ambiguities trigger repetitive self-checking in the tail until hitting the token limit.
Because standard SFT selects and supervises trajectories at the endpoint, filtering by final answers (as in rejection fine-tuning, RFT) discards valid teacher prefixes whenever rollouts fail to finish, while full-trace supervision forces the student to imitate repetitive tail loops.
To address both limitations, we propose \textbf{Prefix Rejection Fine-Tuning (\method{})}, a simple and general prefix-supervised SFT framework built on two core designs.
First, \textbf{Prefix Reward} probes cached teacher prefix states in a single step to rescue unfinished trajectories whose early deduction already points to the correct answer.
Second, \textbf{Prefix Credit Assignment} supervises only the informative reasoning prefix, removes the repetitive tail, and appends a clean stop-and-answer transition.
Across student scales, in-domain and out-of-domain scientific VQA splits, and zero-shot external chart and math benchmarks, \method{} improves student natural accuracy over standard RFT by $+6.7\%$ on Qwen3.5-4B ($68.0\% \to 74.7\%$) and $+4.7\%$ on Qwen3.5-9B ($70.8\% \to 75.5\%$) while cutting thinking tokens by roughly half ($1.97\times$--$2.25\times$), outperforming existing SFT variants and online reinforcement learning baselines.
\end{abstract}

\begin{figure}[t]
  \centering
  \includegraphics[width=\linewidth]{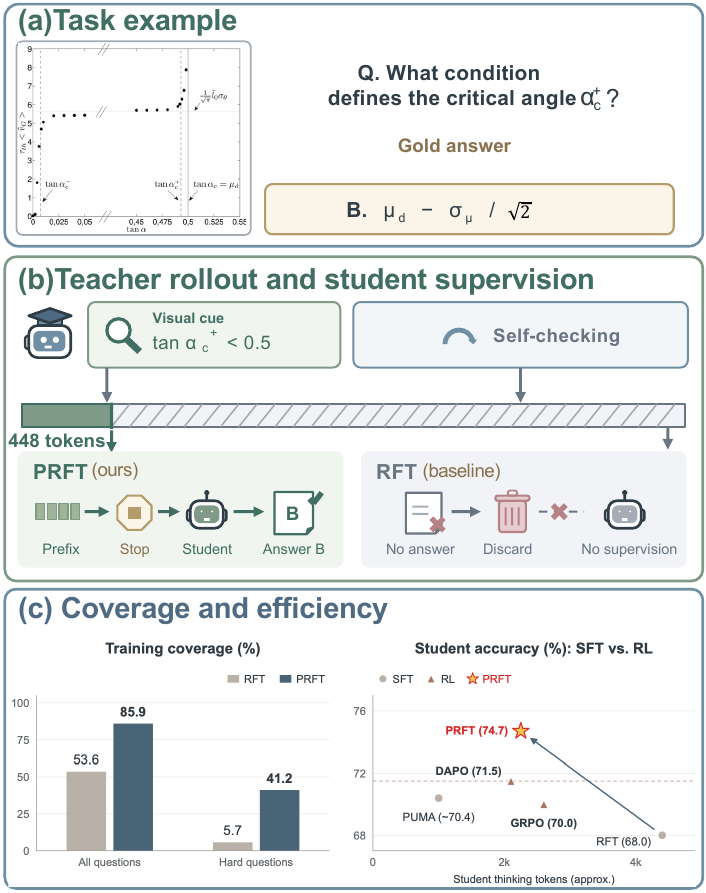}
  \vspace{-3.5mm}
  \caption{\textbf{Standard SFT/RFT~\cite{rftscaling2023,tong2024dart} waits for the ending, whereas \method{} reads the prefix.}
    \emph{(a, b)} On a real teacher rollout drawn to scale, the teacher captures the key visual cues ($\tan\alpha_{\mathrm{c}}^{+} < 0.5$) within the early reasoning prefix ($448$ tokens) before entering $2{,}880$ tokens of repetitive self-checking that hit the token limit without answering. Standard RFT discards this rollout, whereas \method{} probes the cached prefix in one step to recover the answer and trains the student to stop cleanly.
    \emph{(c)} \method{} expands teacher pass rate to $85.9\%$ (on the $9{,}702$-question pool) and improves 4B student accuracy by $+6.7\%$ over RFT while cutting thinking tokens by nearly half.}
  \label{fig:overview}
  \vspace{-3.5mm}
\end{figure}

\section{Introduction}
\label{sec:intro}

Transferring multi-step reasoning from a strong teacher to a student vision-language model (VLM)~\cite{liu2023,internvl2024,bai2025} is a central goal of multimodal post-training~\cite{zhang2023,llavacot2025,rftscaling2023,tong2024dart,zelikman2022star,hu2024visual,bousselham2026vold,yue2026omnithoughtvis,neogi2025docval}.
To achieve this, standard supervised fine-tuning (SFT) pipelines sample teacher trajectories over image-question pairs, apply rejection fine-tuning (RFT) to keep those that finish with the gold answer, and train the student on the surviving traces.
Yet on complex scientific figures~\cite{chartqa2022,mathvista2024,mmmu2024,tang2026vectorharness,mao2026residual,charxiv2024,wang2024,tang2026charts,mao2026textground4m}, this pipeline breaks down (\cref{fig:overview}): teacher rollouts frequently hit the token limit before producing a final answer, causing endpoint filtering to discard nearly half of all training questions (and over $94\%$ of hard questions) while training the student to generate overly long, low-accuracy responses~\cite{chen2025}.

Why do teacher trajectories fail to finish on complex figures?
As shown in \cref{fig:overview}a,b, multimodal reasoning separates early visual perception from late text termination~\cite{yi2018nsvqa,hudson2019gqa}: the teacher identifies key visual cues and encodes the answer evidence within a concise reasoning prefix, but dense axes, overlapping curves, or subtle symbols trap the model in repetitive tail verification until the token budget runs out.
Because existing SFT methods evaluate and supervise trajectories at the endpoint, they face a dilemma between losing valid prefixes and imitating tail loops.
\emph{Unfiltered SFT} trains on all teacher rollouts and exposes the student to both wrong deductions and repetitive tails, whereas \emph{filtered SFT} (such as RFT) discards valid teacher prefixes whenever rollouts fail to finish and still trains the student to copy repetitive tails on surviving traces.

To fully utilize teacher trajectories, SFT must select and supervise reasoning at the prefix rather than the endpoint, retaining valid early deductions while removing repetitive tails.
Motivated by this insight, we propose \textbf{Prefix Rejection Fine-Tuning (\method{})}, a prefix-supervised SFT framework that combines two core designs (\cref{fig:overview,fig:framework}).
For trajectory selection, \textbf{Prefix Reward} branches cached teacher prefix states to check in a single step whether early reasoning has already resolved the answer, rescuing unfinished trajectories that RFT discards.
For token supervision, \textbf{Prefix Credit Assignment} restricts training loss to the informative reasoning prefix, masks out the repetitive tail, and appends a clean stop-and-answer transition so the student learns to reason concisely and stop cleanly.

We analyze prefix pass rates across four teacher models (\textsc{Qwen3.5-9B/27B/35B-A3B}~\cite{qwen35} and \textsc{InternVL2-26B}~\cite{internvl2024,chen2024}) and evaluate \textsc{Qwen3.5-4B} and \textsc{9B} student distillation across \textsc{VeriSciQA} evaluation splits~\cite{verisciqa2026} and zero-shot external benchmarks (\textsc{MathVista}~\cite{mathvista2024}, \textsc{ChartQA}~\cite{chartqa2022}, and \textsc{MathVision}~\cite{wang2024}).
\method{} improves student natural accuracy over standard RFT by $+6.7\%$ on \textsc{Qwen3.5-4B} ($68.0\% \to 74.7\%$) and $+4.7\%$ on \textsc{Qwen3.5-9B} ($70.8\% \to 75.5\%$) while cutting thinking tokens by roughly half ($1.97\times$--$2.25\times$), consistently surpassing unfiltered SFT, filtered SFT variants, and tested online reinforcement learning (RL) baselines such as \textsc{GRPO}~\cite{shao2024deepseekmath,deepseekr12025} and \textsc{DAPO}~\cite{dapo2025}.
In summary, our main contributions are:
\begin{itemize}[leftmargin=*,itemsep=0.8pt,topsep=1.0pt]
  \item \textbf{Challenge of Multimodal Reasoning Distillation:} We reveal that visual overthinking decouples early prefix visual deduction from text termination. This explains why existing SFT and RFT pipelines discard valid prefixes on unfinished rollouts and imitate repetitive tails on surviving traces.
  \item \textbf{Prefix Rejection Fine-Tuning (\method{}):} We propose a prefix-supervised SFT framework that couples \textbf{Prefix Reward} with \textbf{Prefix Credit Assignment} to rescue valid teacher prefixes and remove repetitive tails.
  \item \textbf{Broad Effectiveness and Efficiency:} Across 4B and 9B student scales, internal splits, and external chart/math benchmarks, \method{} improves student accuracy by $+6.7\%$ (4B) and $+4.7\%$ (9B) over standard RFT, surpassing tested SFT and RL baselines while cutting thinking tokens by roughly half ($1.97\times$--$2.25\times$).
\end{itemize}

\section{Related Work}
\label{sec:related}

\paragraph{Supervised fine-tuning, RFT, and self-distillation.}
Supervised fine-tuning (SFT) is the most direct approach to adapt multimodal models~\cite{li2021albef,liu2023,internvl2024,chen2024,bai2025,xue2026distill} for visual and scientific reasoning~\cite{hudson2019gqa,chartqa2022,mathvista2024,zhang2023,mmmu2024,charxiv2024,wang2024}.
Training on answer labels without reasoning (Direct-Answer SFT) invites unimodal language priors and option shortcuts~\cite{goyal2017vqa,agrawal2018vqacp,cadene2019rubi,llavacot2025}, whereas unfiltered CoT SFT~\cite{zhang2023,llavacot2025} trains on all teacher rollouts including wrong deductions.
Filtered SFT methods such as rejection fine-tuning (RFT)~\cite{rftscaling2023}, self-distillation (STaR/SDFT)~\cite{zelikman2022star}, and DART-Math~\cite{tong2024dart} filter rollouts by terminal answer correctness, but break down when complex figures~\cite{charxiv2024,verisciqa2026} trigger non-terminating verification loops.
\method{} extends RFT by probing the teacher's cached prefix in one step to rescue unfinished rollouts.

\paragraph{Online reinforcement learning and logit distillation.}
Starting from the same base model, on-policy RL (\textsc{GRPO}~\cite{shao2024deepseekmath,deepseekr12025}, \textsc{DAPO}~\cite{dapo2025}) optimizes group-relative student rewards, length-penalized RL~\cite{kimi2025k15} and \textsc{S-GRPO}~\cite{sgrpo2025} penalize verbosity or encourage early exits, whereas logit distillation (\textsc{OPD}~\cite{agarwal2023,minillm2024,yang2026oprd}) matches teacher output distributions, with \textsc{VOLD}~\cite{bousselham2026vold} extending OPD to multimodal reasoning.
On complex figures, however, online RL loses learning signal when student rollouts fail on hard items, \textsc{OPD} requires caching teacher top-$k$ logits, and early-exit SFT (\textsc{PUMA}~\cite{puma2026}, \textsc{Settle}~\cite{settle2026}) cuts off intermediate deduction.
\method{} retains the full deduction prefix while removing repetitive tail loops, surpassing both SFT and RL configurations under natural termination.

\paragraph{Test-time budget control and early stopping.}
To mitigate overthinking~\cite{chen2025}, forcing a model to emit an answer when generation reaches a token limit or confidence threshold is a common test-time heuristic~\cite{s12025,budgetguidance2025,truncbudget2026}, yet it leaves verbose generation habits unchanged during deployment.
Rather than relying on external test-time intervention, \method{} trains the student to reason concisely and stop cleanly during SFT while preserving prefix deduction accuracy.

\section{Methodology}
\label{sec:method}

\begin{figure*}[t]
  \centering
  \includegraphics[width=\textwidth]{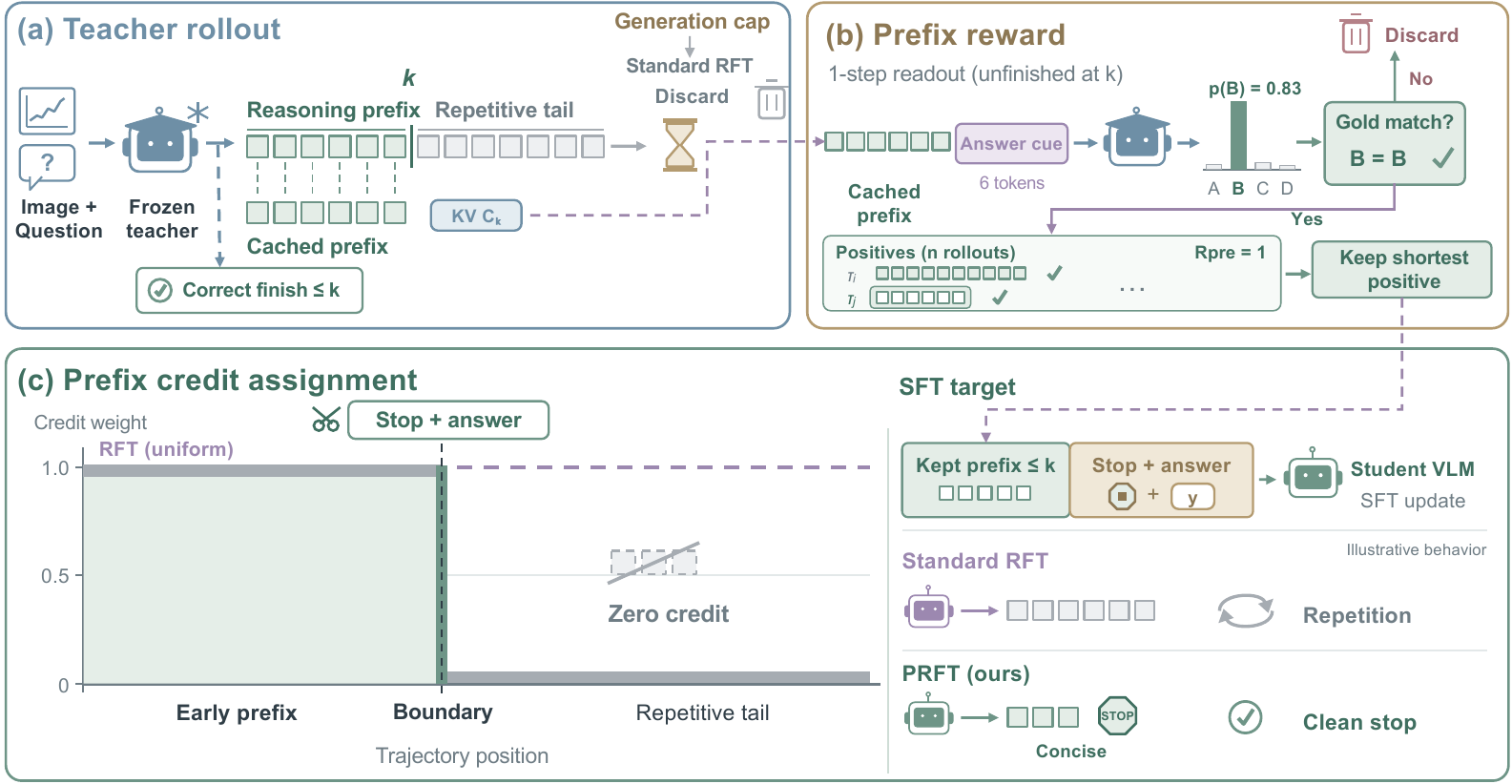}
  \vspace{-4.5mm}
  \caption{\textbf{\method{} framework.}
    \emph{(a) Teacher rollout:} A frozen teacher generates each rollout once and caches its prefix states (reasoning prefix up to boundary $k$, followed by a repetitive tail that hits the generation cap without answering).
    \emph{(b) Prefix reward:} For rollouts unfinished at $k$, \method{} branches the cached prefix states with a short answer cue to decode the teacher's option distribution in a single step ($<1\%$ online warm-cache compute overhead) and retains the shortest positive prefix.
    \emph{(c) Prefix credit assignment:} \method{} assigns uniform positive credit ($1.0$) to prefix tokens ($t \le k$) and zero credit to the repetitive tail ($t > k$, unlike full-trace Standard RFT, dashed; optional visual-gain token weighting is ablated in \cref{tab:ablation}a), and appends a clean stop-and-answer transition so the student learns concise, self-terminating reasoning.}
  \label{fig:framework}
  \vspace{-4mm}
\end{figure*}

\textbf{Prefix Rejection Fine-Tuning (\method{})} is a prefix-supervised SFT framework that shifts both trajectory selection and token supervision from the trajectory ending to the reasoning prefix (\cref{fig:framework}).
We first analyze why endpoint-based SFT and RFT break down on complex visual inputs (\cref{sec:diagnosis}), and then present the unified \method{} framework (\cref{sec:readout}).

\subsection{Motivating Observations}
\label{sec:diagnosis}

\paragraph{Visual deduction finishes early while the tail loops.}
When a multimodal teacher solves a complex scientific figure, visual deduction and text termination are decoupled (\cref{fig:overview}).
The teacher's reliance on the image is concentrated in the early reasoning prefix and drops sharply toward the tail (\cref{fig:vis-decay}a).
Once the teacher encodes the key visual cues in the prefix, fine-grained visual ambiguities such as overlapping curves or dense axis ticks frequently trigger repetitive self-checking loops until hitting the token limit.

\paragraph{Why endpoint-based post-training struggles.}
Evaluating and supervising trajectories at the endpoint therefore creates a clear dilemma: \emph{unfiltered SFT} forces the student to imitate both wrong deductions and tail loops, whereas \emph{filtered SFT} (RFT~\cite{rftscaling2023,tong2024dart}) discards valid prefixes whenever rollouts fail to finish and still teaches tail looping on surviving traces.
Meanwhile, truncating at the very first correct guess~\cite{puma2026,settle2026} cuts off essential intermediate deduction steps, and student-only online RL (\textsc{GRPO}~\cite{shao2024deepseekmath}, \textsc{DAPO}~\cite{dapo2025}) loses learning signal when the student cannot solve hard figures on its own.
Effective distillation thus requires rescuing valid teacher prefixes while supervising only the informative deduction steps.

\subsection{\method{} Framework}
\label{sec:readout}
\label{sec:solver}

\begin{table*}[t]
  \centering
  \footnotesize
  \setlength{\tabcolsep}{4.2pt}
  \renewcommand{\arraystretch}{1.00}
  \resizebox{\textwidth}{!}{%
  \begin{tabular}{l l l c c c c c}
    \toprule
    \textbf{Method}
    & \textbf{Trajectory Selection}
    & \textbf{Token Supervision}
    & \textbf{Acc.\ (\%)}
    & \textbf{Maj@4 (\%)}
    & \textbf{Finish (\%)}
    & \textbf{Think Tokens}
    & \textbf{Token Change}
    \\
    \midrule
    \rowcolor{fam}[\tabcolsep][\tabcolsep]
    \multicolumn{8}{c}{\textit{Qwen3.5-4B --- Untrained Controls}}
    \\
    Untrained Base & -- & -- & 61.0 & 71.1 & 74.9 & 5,649 & \textcolor{qlbverm}{$\uparrow$ 27.2\%} \\
    Forced Cap Readout~\cite{budgetguidance2025,truncbudget2026} & -- & -- & 73.2 & 76.8 & 74.9$^\ast$ & 5,649 & \textcolor{qlbverm}{$\uparrow$ 27.2\%} \\
    Confidence Early Exit & -- & -- & 73.2 & 76.5 & -- & 3,988 & \textcolor{qlbgreen!80!black}{$\downarrow$ 10.2\%} \\
    \midrule
    \rowcolor{fam}[\tabcolsep][\tabcolsep]
    \multicolumn{8}{c}{\textit{Qwen3.5-4B --- SFT Family}}
    \\
    Direct-Answer SFT & RFT pool (6,656) & Answer label only & 71.0 & 71.4 & 100.0 & 0 & \textcolor{qlbgreen!80!black}{$\downarrow$ 100\%} \\
    Standard RFT~\cite{rftscaling2023} & Natural endpoint (6,656) & Full trace (uniform) & 68.0 & 73.5 & 86.4 & 4,441 & Baseline \\
    Student Self-Distillation (\textsc{SDFT}~\cite{zelikman2022star})$^\dagger$ & Student endpoint & Full student trace & 71.5 & -- & 99.5 & 1,740 & \textcolor{qlbgreen!80!black}{$\downarrow$ 60.8\%} \\
    Top-16 Logit KL (\textsc{OPD}~\cite{agarwal2023,minillm2024}) & Teacher prefixes (9,366) & Top-16 forward KL & 72.3 & -- & 98.9 & 2,254 & \textcolor{qlbgreen!80!black}{$\downarrow$ 49.2\%} \\
    \textsc{PUMA}~\cite{puma2026} & First correct probe (9,366) & Early prefix ($714$ tgt) & 70.4 & -- & 99.9 & 1,070 & \textcolor{qlbgreen!80!black}{$\downarrow$ 75.9\%} \\
    \textsc{Settle}~\cite{settle2026} & First stable probe (9,366) & Early prefix ($739$ tgt) & 70.9 & 72.8 & 99.9 & 1,032 & \textcolor{qlbgreen!80!black}{$\downarrow$ 76.8\%} \\
    \midrule
    \rowcolor{fam}[\tabcolsep][\tabcolsep]
    \multicolumn{8}{c}{\textit{Qwen3.5-4B --- RL Family}}
    \\
    \textsc{GRPO}~\cite{shao2024deepseekmath}$^\dagger$ & Student group (12 rds) & Group advantage & 70.0 & -- & 96.5 & 2,564 & \textcolor{qlbgreen!80!black}{$\downarrow$ 42.3\%} \\
    Length-Penalty RL~\cite{kimi2025k15} & Student group & Advantage + length pen. & 69.7 & -- & 99.9 & 1,353 & \textcolor{qlbgreen!80!black}{$\downarrow$ 69.5\%} \\
    \textsc{DAPO}~\cite{dapo2025} & Dynamic group (6 rds) & Group advantage & 71.5 & 74.9 & 98.4 & 2,084 & \textcolor{qlbgreen!80!black}{$\downarrow$ 53.1\%} \\
    \textsc{S-GRPO}~\cite{sgrpo2025} & Student group & Early-exit reward & 66.4 & 73.1 & 84.2 & 4,582 & \textcolor{qlbverm}{$\uparrow$ 3.2\%} \\
    \midrule
    \rowcolor{fam}[\tabcolsep][\tabcolsep]
    \multicolumn{8}{c}{\textit{Qwen3.5-4B --- \method{} (Ours)}}
    \\
    \rowcolor{ours}[\tabcolsep][\tabcolsep]
    \method{} (Matched-Row, Step Cut) & Prefix Reward (6,656) & Prefix Credit & 73.7 & \textbf{76.8} & 98.8 & \textbf{2,188} & \textcolor{qlbgreen!80!black}{\textbf{$\downarrow$ 50.7\%}} \\
    \rowcolor{ours}[\tabcolsep][\tabcolsep]
    \textbf{\method{} (Shared-Pool, Budget Cut)} & \textbf{Prefix Reward (7,621)} & \textbf{Prefix Credit} & \textbf{74.7} & 76.2 & \textbf{100.0} & 2,257 & \textcolor{qlbgreen!80!black}{$\downarrow$ 49.2\%} \\
    \midrule
    \rowcolor{fam}[\tabcolsep][\tabcolsep]
    \multicolumn{8}{c}{\textit{Qwen3.5-9B --- Scale-Up}}
    \\
    Untrained Base (9B) & -- & -- & 65.4 & 74.8 & 79.2 & 5,171 & \textcolor{qlbverm}{$\uparrow$ 18.7\%} \\
    Standard RFT (9B)~\cite{rftscaling2023} & Natural endpoint (6,656) & Full trace (uniform) & 70.8 & 76.7 & 87.4 & 4,355 & Baseline \\
    \rowcolor{ours}[\tabcolsep][\tabcolsep]
    \textbf{\method{} (Matched-Row, 9B)} & \textbf{Prefix Reward (6,656)} & \textbf{Prefix Credit} & \textbf{75.5} & \textbf{77.4} & \textbf{99.8} & \textbf{1,934} & \textcolor{qlbgreen!80!black}{\textbf{$\downarrow$ 55.6\%}} \\
    \midrule
    \rowcolor{fam}[\tabcolsep][\tabcolsep]
    \multicolumn{8}{c}{\textit{Teacher Reference (1 decode seed)}}
    \\
    Teacher Ref. / \textsc{Qwen3.5-27B} & -- & -- & 72.5 & -- & 84.3 & 4,602 & \textcolor{qlbverm}{$\uparrow$ 3.6\%} \\
    \bottomrule
  \end{tabular}%
  }
  \vspace{-1.5mm}
  \caption{\textbf{Main comparison across SFT and RL post-training paradigms at 4B and 9B scales on \textsc{VeriSciQA} Main ($1{,}200$ items).}
    Reported metrics are single-sample natural accuracy (\textbf{Acc.}, $4$ decoding seeds unless marked $^\dagger$ for $2$ seeds), 4-sample majority voting accuracy (\textbf{Maj@4}, `--' where $4$-seed votes are unavailable), natural completion rate within the $12{,}000$-token cap (\textbf{Finish}; $^\ast$Forced Cap Readout shares Base's $74.9\%$ natural finish and reaches $99.0\%$ answer availability via forced cap readout), and effective thinking tokens (training scopes and forced-cap comparison in \cref{app:impl,app:matched}).
    Standard RFT lags behind on-policy RL (\textsc{GRPO}, \textsc{DAPO}) because tail loops lower completion ($86.4\%$ Finish), whereas \method{} achieves near-$100\%$ completion and surpasses both SFT ($+5.7\%$ Matched-Row, $+6.7\%$ Shared-Pool) and RL ($+3.2\%$ to $+4.7\%$) while cutting thinking tokens by roughly half.}
  \label{tab:main-results}
  \vspace{-4mm}
\end{table*}

\paragraph{Problem formulation and cached prefix readout.}
Given an image-question input $x$ and gold answer $y$, a teacher model generates a reasoning trajectory $\tau = (t_1, \dots, t_L)$, where $\tau_{\le k} = (t_1, \dots, t_k)$ denotes the reasoning prefix up to boundary $k$.
If $\tau$ finishes naturally within $k$ tokens, we directly use its output answer as $\hat{y}_k$; otherwise, we branch the teacher's cached key-value states at prefix $\tau_{\le k}$, append a short answer cue, and decode the teacher's predicted answer $\hat{y}_k$ in a single forward pass (\cref{fig:framework}):
\begin{equation}
  \hat{y}_k = \operatorname{Readout}(x, \tau_{\le k}).
  \label{eq:ir}
\end{equation}
Because the prefix states of $\tau_{\le k}$ are already computed during trajectory generation, this single-step readout adds negligible compute under prefix caching and directly connects trajectory selection with token supervision.

\paragraph{Design 1: Prefix Reward for trajectory selection.}
Instead of waiting for natural termination at the end of the trajectory, we select training trajectories using the \textbf{Prefix Reward} $R_{\mathrm{pre}}(\tau, y) \in \{0, 1\}$:
\begin{equation}
  R_{\mathrm{pre}}(\tau, y) = \mathbf{1}\bigl[\hat{y}_k = y\bigr],
  \label{eq:prefix_reward}
\end{equation}
which equals $1$ whenever the prefix $\tau_{\le k}$ has already reached the gold answer $y$ (either by finishing naturally within $k$ tokens or via the single-step readout $\hat{y}_k$ in \cref{eq:ir} passing the option-renormalized confidence filter), and $0$ otherwise.
When multiple rollouts per question achieve $R_{\mathrm{pre}} = 1$, we select the shortest positive prefix.
By evaluating correctness at the prefix rather than the endpoint, Prefix Reward rescues unfinished teacher deductions on hard visual problems without extra sampling.

\paragraph{Design 2: Prefix Credit Assignment for token supervision.}
To prevent the student model $P_\theta$ (with trainable parameters $\theta$) from imitating repetitive tail loops, \textbf{Prefix Credit Assignment} assigns uniform positive weight $w_t = 1$ to tokens within the prefix ($t \le k$; optional visual-gain weighting $w_t \propto 1 + \max(\Delta_t, 0)$ is ablated in \cref{tab:ablation}a) and assigns zero weight to all tail tokens ($w_t = 0$ for $t > k$; \cref{fig:framework}c).
Removing the repetitive tail beyond $k$ and appending a clean stop-and-answer transition for $y$ yields the unified \method{} SFT loss:
\begin{equation}
  \begin{aligned}
    \mathcal{L}_{\mathrm{\method{}}}(\theta) = - R_{\mathrm{pre}}(\tau, y) \Biggl(
    &\sum_{t=1}^{k} w_t \log P_\theta(t_t \mid x, t_{<t}) \\[-1.5mm]
    &\!\!+ \log P_\theta(y \mid x, \tau_{\le k})\Biggr).
  \end{aligned}
  \label{eq:admit_gate}
\end{equation}
Here the first term supervises only the informative prefix tokens $t_1, \dots, t_k$, while the second term trains the student to stop and output the answer $y$ right after $\tau_{\le k}$, turning unfinished rollouts into concise targets.

\paragraph{Offline prefix boundary rules and task generality.}
The prefix boundary $k$ balances complete deduction against concise termination: cutting too early removes intermediate steps, whereas cutting too late retains repetitive tail checking.
Because prefix readout (\cref{eq:ir}) leaves $\tau$ intact and Prefix Credit Assignment (\cref{eq:admit_gate}) masks tokens after $k$, we evaluate two simple deterministic boundary rules offline without extra teacher rollouts: \textbf{Step Cut} (the latest paragraph break within $[256, 3{,}072]$) and \textbf{Budget Cut} (the full prefix up to $k = 3{,}072$ tokens, which also retains traces lacking early paragraph breaks; see \cref{app:impl}).
Relying only on standard autoregressive key-value states rather than task-specific priors, \method{} applies cleanly across model scales and general multimodal benchmarks.

\section{Experiments}
\label{sec:exp}

We organize our evaluation into four parts:
(1) main comparison against existing SFT variants and online RL across 4B and 9B student models (\cref{sec:exp:acc});
(2) generalization across unseen domains, external benchmarks, and input modalities (\cref{sec:exp:grounded});
(3) ablation studies on \method{} components, prefix length, test-time token budgets, and teacher sampling (\cref{sec:exp:ablation}); and
(4) qualitative visualizations and visual reliance analysis (\cref{sec:exp:mechanism}).

\subsection{Experimental Setup}
\label{sec:setup}

\paragraph{Datasets and models.}
We evaluate on \textsc{VeriSciQA}~\cite{verisciqa2026} using an $11{,}702$-question training pool ($9{,}702$ sampled at a $3{,}328$-token limit and $2{,}000$ at $8{,}192$), a $1{,}627$-question Hard-Tail Pool ($n{=}8$), four paper-disjoint evaluation splits (Main $1{,}200$, Held-Out $1{,}379$ with $1{,}053$ visually dependent items, Unseen Reserve $1{,}216$ spanning 20 STEM subjects and 11 chart types, and OOD Caption-Free Visual $369$), and a $2{,}306$-question teacher-rejected Hard-Tail diagnostic slice ($0$ overlap with Standard RFT, Prefix Credit Only, Step Cut, and Full Pool SFT), alongside zero-shot transfer to \textsc{MathVista}~\cite{mathvista2024}, \textsc{ChartQA}~\cite{chartqa2022}, and \textsc{MathVision}~\cite{wang2024}.
We analyze prefix pass rates across four teacher models (\textsc{Qwen3.5-9B}, \textsc{27B}, \textsc{35B-A3B}~\cite{qwen35}, and \textsc{InternVL2-26B}~\cite{internvl2024,chen2024}) and evaluate \textsc{Qwen3.5-27B} distillation into \textsc{Qwen3.5-4B} and \textsc{9B} students (training status of cross-family \textsc{InternVL2-4B}/\textsc{Phi-3}~\cite{chen2024,phi3technicalreport2024} checkpoints is summarized in \cref{app:impl}).
All students are trained with LoRA~\cite{lora2022} ($r{=}64, \alpha{=}128$, lr $10^{-4}$, batch size $32$) for three epochs from the same base checkpoint at a $3{,}072$-token prefix budget on NVIDIA H100 GPUs (\cref{app:impl}).

\paragraph{Baselines and evaluation metrics.}
We compare four method families from the same base checkpoint:
(i) \textbf{Untrained Controls:} Untrained Base, Forced Cap Readout~\cite{budgetguidance2025,truncbudget2026}, and Confidence Early Exit;
(ii) \textbf{SFT Family:} Direct-Answer SFT, Standard RFT~\cite{rftscaling2023}, \textsc{SDFT}~\cite{zelikman2022star}, Top-16 Logit KL (\textsc{OPD}~\cite{agarwal2023,minillm2024}), and Early-Exit SFT (\textsc{PUMA}~\cite{puma2026}, \textsc{Settle}~\cite{settle2026});
(iii) \textbf{RL Family:} \textsc{GRPO}~\cite{shao2024deepseekmath,deepseekr12025}, Length-Penalty RL~\cite{kimi2025k15}, \textsc{DAPO}~\cite{dapo2025}, and \textsc{S-GRPO}~\cite{sgrpo2025}; and
(iv) \textbf{\method{} (Ours):} Matched-Row Step Cut ($6{,}656$ rows, $624$ steps) and Shared-Pool Budget Cut ($7{,}621$ rows, $714$ steps).
We report single-sample natural \textbf{Accuracy (\%)} (averaging $4$ decoding seeds under a $12{,}000$-token limit, scoring parsed post-\texttt{</think>} answers), 4-sample majority voting \textbf{Maj@4 (\%)}~\cite{selfconsistency2023}, natural completion rate \textbf{Finish (\%)} (responses stopping before the token limit), effective \textbf{Thinking Tokens}, and teacher \textbf{Pass Rate (\%)} (questions yielding a valid training target).

\subsection{Main Comparison}
\label{sec:exp:acc}

\begin{figure*}[t]
  \centering
  \includegraphics[width=\textwidth]{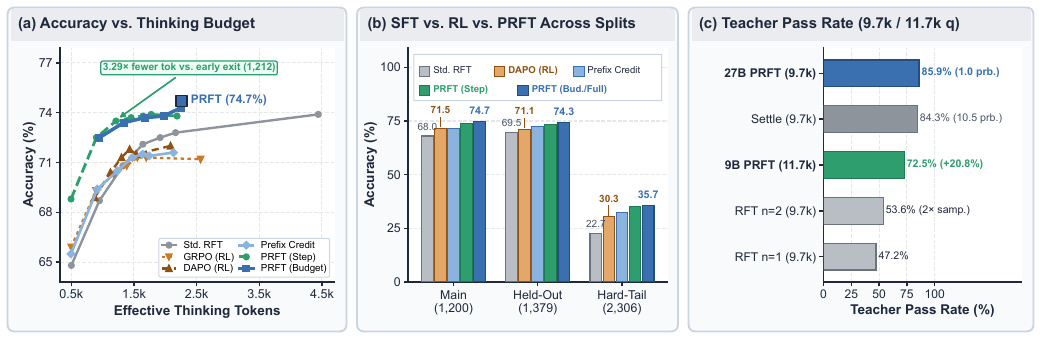}
  \vspace{-4.5mm}
  \caption{\textbf{Accuracy-token frontier, SFT vs.\ RL comparison across splits, and teacher pass rate.}
    \emph{(a)} Across verified test-time thinking budgets on Main ($1{,}200$ items), \method{} outperforms tested SFT and RL baselines, matching the strongest untrained confidence early-exit baseline ($73.2\%$ at $3{,}988$ tok) in $1{,}212$ thinking tokens ($3.29\times$ fewer).
    \emph{(b)} Across Main, Held-Out, and Hard-Tail splits, Standard RFT lags behind RL (\textsc{DAPO}), whereas both Matched-Row Step Cut and Budget/Full-Pool \method{} surpass both paradigms.
    \emph{(c)} Single-step prefix probing expands teacher pass rate on both $9{,}702$-question (\textsc{27B}) and $11{,}702$-question (\textsc{9B}) pools without extra sampling.}
  \label{fig:exp_summary}
  \vspace{-4mm}
\end{figure*}

\paragraph{Standard SFT suffers from discarded prefixes and tail loops.}
In \cref{tab:main-results}, all post-trained 4B models start from the same \textsc{Qwen3.5-4B} checkpoint.
Training directly on answer labels without reasoning (Direct-Answer SFT) encourages the student to exploit non-visual priors and option shortcuts~\cite{agrawal2018vqacp,cadene2019rubi}.
Conversely, Standard RFT ($68.0\%$) supervises full teacher traces and teaches the student to imitate repetitive tail checking (\cref{fig:case-studies}b): forcing the Standard RFT student to output an answer at the $12{,}000$-token limit raises its accuracy to $73.9\%$ (\cref{tab:multiseed-cap}b), showing that the student often resolves the deduction early but fails to terminate.
Other SFT variants mitigate looping with clear trade-offs: self-distillation (\textsc{SDFT}) is bottlenecked by the student's own rollout capacity, early-exit methods (\textsc{PUMA}, \textsc{Settle}) cut off intermediate deduction steps before visual reasoning finishes, and top-$16$ logit distillation (\textsc{OPD}) requires caching teacher top-$k$ logits ($1.47\text{ GiB}$).
Existing SFT pipelines thus struggle to preserve complete teacher deductions without copying repetitive tails.

\paragraph{Online RL reduces looping but plateaus on hard figures.}
On-policy RL methods such as \textsc{GRPO} ($70.0\%$) and \textsc{DAPO} ($71.5\%$) outperform Standard RFT in \cref{tab:main-results} because penalizing unfinished rollouts teaches the student to stop before hitting the token limit ($96.5\%$--$98.4\%$ Finish).
However, online RL optimizes only the student's self-sampled trajectories.
On complex figures that exceed the student's initial capability, student rollouts fail and group-relative advantages vanish, limiting visual deduction gains.
Adding explicit length penalties ($69.7\%$) or early-exit rewards (\textsc{S-GRPO}, $66.4\%$ natural / $73.6\%$ forced cap) adjusts generation length without injecting new reasoning demonstrations on hard items.

\paragraph{\method{} outperforms tested SFT and RL at half the tokens.}
\method{} resolves both limitations within standard SFT (\cref{tab:main-results}).
At strictly matched training rows and steps ($6{,}656$ questions, $624$ updates), \method{} (Matched-Row, Step Cut) reaches \textbf{73.7\%} natural accuracy (\textbf{76.8\%} Maj@4, \textbf{+5.7\%} over Standard RFT) while cutting thinking tokens by \textbf{50.7\%} ($2{,}188$ vs.\ $4{,}441$).
Retaining traces without early paragraph breaks in \method{} (Shared-Pool, Budget Cut; $7{,}621$ rows, $714$ steps) further reaches \textbf{74.7\%} accuracy (\textbf{+6.7\%} over Standard RFT and \textbf{+3.2\%} over \textsc{DAPO}) with $100.0\%$ natural completion and $49.2\%$ fewer thinking tokens ($2{,}257$, $1.97\times$ reduction).
Because the student learns early visual deduction without inheriting tail loops, the 4B student surpasses the 27B teacher under natural termination ($72.5\%$, $84.3\%$ Finish; note forced cap readout on the 27B teacher reaches $81.3\%$).

\paragraph{\method{} scales to 9B students and speeds up inference.}
The same natural-decoding trend holds when scaling to \textsc{Qwen3.5-9B} (\cref{tab:main-results}, Matched-Row $6{,}656$ rows), where \method{} reaches \textbf{75.5\%} accuracy (\textbf{+4.7\%} over Standard RFT, \textbf{77.4\%} Maj@4) while cutting thinking tokens by $2.25\times$ ($4{,}355 \to 1{,}934$, $99.8\%$ Finish).
As detailed in \cref{app:matched}, forced cap readout narrows or reverses the gap on Held-Out and 9B Standard RFT ($76.7\%$ Cap at $4{,}355$ tok), confirming that \method{}'s core gain is internalizing concise, self-terminating reasoning without external cap intervention.
Halving output tokens yields measured $1.14\times$--$1.25\times$ wall-clock speedups on $4\times\text{H100}$ GPUs (\cref{app:latency}).

\subsection{Generalization and Modality Study}
\label{sec:exp:grounded}

\begin{table}[t]
  \centering
  \footnotesize
  \setlength{\tabcolsep}{2.2pt}
  \renewcommand{\arraystretch}{0.98}
  \resizebox{\linewidth}{!}{%
  \begin{tabular}{lccccccc}
    \toprule
    \rowcolor{fam}[\tabcolsep][\tabcolsep]
    \textbf{(a) Split / Benchmark}
    & \textbf{Held-Out}
    & \textbf{H.-O. Vis.}
    & \textbf{Reserve$^\dagger$}
    & \textbf{OOD Vis.}
    & \textbf{Hard-Tail}
    & \textbf{MathVista}
    & \textbf{ChartQA}
    \\
    \midrule
    Standard RFT (SFT) & 69.5 & -- & 69.4 & 85.5 & 22.7 & 78.1 & 59.7 \\
    On-Policy RL (\textsc{DAPO}) & 71.1 & 67.7 & -- & -- & 30.3 & 75.0 & 59.6 \\
    Prefix Credit Only & 72.5 & 68.6 & 72.6 & 85.6 & 32.2 & -- & -- \\
    \rowcolor{ours}[\tabcolsep][\tabcolsep]
    \method{} (Step Cut) & 73.2 & -- & \textbf{73.2} & 86.2 & 35.3 & 79.2 & 60.4 \\
    \rowcolor{ours}[\tabcolsep][\tabcolsep]
    \textbf{\method{} (Budget / Full$^\ast$)} & \textbf{74.3}$^\ast$ & \textbf{70.4}$^\ast$ & -- & \textbf{86.6}$^\ast$ & \textbf{35.7}$^\ast$ & \textbf{81.8} & \textbf{68.6} \\
    \midrule
    \rowcolor{fam}[\tabcolsep][\tabcolsep]
    \textbf{(b) Input Modality (Main)}
    & \multicolumn{2}{c}{\textbf{Image + Cap.}}
    & \textbf{Image Only}
    & \textbf{Caption Only}
    & \textbf{Neither}
    & \multicolumn{2}{c}{\textbf{Img.\ Drop}}
    \\
    \midrule
    Untrained Base & \multicolumn{2}{c}{61.0} & 57.9 & 35.4 & 32.6 & \multicolumn{2}{c}{$-25.6\%$} \\
    On-Policy RL (\textsc{DAPO}) & \multicolumn{2}{c}{71.5} & 68.3 & 36.8 & 33.4 & \multicolumn{2}{c}{$-34.7\%$} \\
    Prefix Credit Only & \multicolumn{2}{c}{71.3} & 68.8 & 39.8 & 36.1 & \multicolumn{2}{c}{$-31.5\%$} \\
    \rowcolor{ours}[\tabcolsep][\tabcolsep]
    \method{} (Full Pool) & \multicolumn{2}{c}{73.2} & \textbf{72.0} & 40.3 & -- & \multicolumn{2}{c}{$-32.9\%$} \\
    \rowcolor{ours}[\tabcolsep][\tabcolsep]
    \textbf{\method{} (Step Cut)} & \multicolumn{2}{c}{73.7} & -- & 42.7 & 37.1 & \multicolumn{2}{c}{$-31.0\%$} \\
    Answer-Only SFT (w/o Prefix) & \multicolumn{2}{c}{74.8} & 71.1 & 53.2 & \textcolor{qlbverm}{\textbf{50.4}} & \multicolumn{2}{c}{$-21.6\%$} \\
    \bottomrule
  \end{tabular}%
  }
  \vspace{-1.8mm}
  \caption{\textbf{Generalization across internal splits, external benchmarks, and input modalities (\textsc{Qwen3.5-4B}, Accuracy \%).}
    \emph{(a)} Verified results across Held-Out ($1{,}379$), Held-Out Visually Dependent ($1{,}053$), Unseen Reserve ($1{,}216$; $^\dagger$3-seed mean), OOD Caption-Free ($369$), teacher-rejected Hard-Tail ($2{,}306$), and zero-shot \textsc{MathVista} ($1{,}000$) and \textsc{ChartQA} ($2{,}500$). Entries marked $^\ast$ correspond to Full-Pool Step Cut; \textsc{MathVista}/\textsc{ChartQA} report Budget Cut (full breakdown with \textsc{MathVision} in \cref{tab:external-benchmarks}).
    \emph{(b)} Removing the figure (\texttt{Caption Only}) causes a $>31\%$ accuracy drop across reasoning models, whereas removing captions (\texttt{Image Only}) drops \method{} (Full Pool) by only $1.2\%$ ($73.2\% \to 72.0\%$).}
  \label{tab:grounded}
  \label{tab:modality-grid}
  \vspace{-3.5mm}
\end{table}

\paragraph{\method{} generalizes to unseen domains and external benchmarks.}
Across verified internal splits and external benchmarks in \cref{tab:grounded}a and \cref{fig:exp_summary}b, \method{} consistently outperforms Standard RFT and \textsc{DAPO}.
On \textsc{VeriSciQA}, gains hold across Held-Out ($73.2\%$--$74.3\%$ vs.\ $69.5\%$ RFT), Unseen Reserve ($73.2\%$ 3-seed mean vs.\ $69.4\%$ RFT), OOD Caption-Free ($86.2\%$--$86.6\%$ vs.\ $85.5\%$ RFT), and the teacher-rejected Hard-Tail slice ($35.3\%$--$35.7\%$ vs.\ $22.7\%$ RFT, \textbf{+12.6\% to +13.0\%}).
In zero-shot evaluation (single decoding seed; protocol and \textsc{MathVision} boundary in \cref{app:grounded}), \method{} (Budget Cut) reaches \textbf{81.8\%} on \textsc{MathVista} and \textbf{68.6\%} on \textsc{ChartQA} with $100\%$ completion while generalizing to open-ended numerical questions ($74.8\%$ on \textsc{MathVista} Free-Form).

\paragraph{\method{} preserves visual reliance rather than shortcut collapse.}
To verify visual dependence, \cref{tab:modality-grid}b ablates image and caption inputs at test time on Main ($1{,}200$ items).
Removing the figure (\texttt{Caption Only}) causes a sharp accuracy drop of $31.0\%$--$32.9\%$ for \method{} (comparable to $31.5\%$ for Prefix Credit Only and $34.7\%$ for \textsc{DAPO}), confirming strong reliance on the figure~\cite{goyal2017vqa,cadene2019rubi}.
Conversely, removing the text caption while keeping the figure (\texttt{Image Only}) barely affects \method{} (Full Pool: $73.2\% \to 72.0\%$), widening its margin over Prefix Credit Only ($68.8\%$) and \textsc{DAPO} ($68.3\%$).
In contrast, training on answer labels without reasoning (\texttt{Answer-Only SFT}) retains $50.4\%$ accuracy when both the figure and caption are removed (\texttt{Neither}), and $49.0\%$ on Hard-Tail with question and options alone (\cref{app:blindchannel}), showing that supervising the reasoning prefix helps preserve visual dependence.

\subsection{Ablation Study}
\label{sec:exp:ablation}
\label{sec:exp:budget}
\label{sec:exp:cov}

\begin{table}[t]
  \centering\footnotesize
  \setlength{\tabcolsep}{3.0pt}
  \renewcommand{\arraystretch}{0.94}
  \resizebox{\linewidth}{!}{%
  \begin{tabular}{@{}lcccc@{}}
    \toprule
    \textbf{(a) Component Ablation} & Acc.\ (\%) & Gain & Think Tokens & Option Bias \\
    \midrule
    Standard RFT (Baseline) & 68.0 & Base & 4,441 & 4.8\% \\
    Unfiltered SFT (All Traces, 1 ep.) & 71.9 & $+3.9\%$ & 2,068 & 5.9\% \\
    Answer-Only SFT (w/o Prefix) & 74.8 & $+6.8\%$ & 0 & \textcolor{qlbverm}{\textbf{25.4\%}} \\
    + Prefix Credit Only (w/o $R_{\mathrm{pre}}$) & 71.3 & $+3.3\%$ & 2,131 & 2.7\% \\
    + Prefix Reward (\method{}, Full Pool) & 73.2 & $+5.2\%$ & \textbf{1,849} & 3.5\% \\
    + Prefix Reward (\method{}, Step Cut) & 73.7 & $+5.7\%$ & 2,188 & 3.2\% \\
    + Visual-Gain Weighting & 74.2 & $+6.2\%$ & 2,308 & 3.4\% \\
    + Vision LoRA & 74.5 & $+6.5\%$ & 2,220 & \textbf{1.6\%} \\
    \textbf{\method{} (Shared-Pool, Budget Cut)} & \textbf{74.7} & \textbf{+6.7\%} & 2,257 & 6.3\% \\
    \midrule
    \textbf{(b) Kept Prefix Length} & Acc.\ (\%) & Gain & Think Tokens & Option Bias \\
    \midrule
    Earliest Correct (\textsc{PUMA}) & 70.4 & $+2.4\%$ & 1,070 & 8.3\% \\
    Earliest Stable (\textsc{Settle}) & 70.9 & $+2.9\%$ & 1,032 & 8.1\% \\
    Random Correct Step & 71.9 & $+3.9\%$ & 1,364 & 4.9\% \\
    Latest Step Boundary (Shared Pool) & 73.2 & $+5.2\%$ & 1,865 & \textbf{2.8\%} \\
    \textbf{Full Budget Prefix (\method{}, Budget Cut)} & \textbf{74.7} & \textbf{+6.7\%} & 2,257 & 6.3\% \\
    \bottomrule
  \end{tabular}%
  }
  \vspace{-1.5mm}
  \caption{\textbf{Ablation of \method{} components and kept prefix length on \textsc{Qwen3.5-4B} Main.}
    \textbf{Option Bias} is the max-minus-min natural accuracy spread across gold options A--D.
    \emph{(a)} Prefix Credit Only removes tail loops to match online RL, while adding Prefix Reward injects rescued hard prefixes to surpass both SFT and RL with low option bias.
    \emph{(b)} Retaining a more complete reasoning prefix monotonically improves student accuracy, whereas stripping the prefix (\texttt{Answer-Only SFT}) induces a $25.4\%$ option bias.}
  \label{tab:ablation}
  \vspace{-2.5mm}
\end{table}

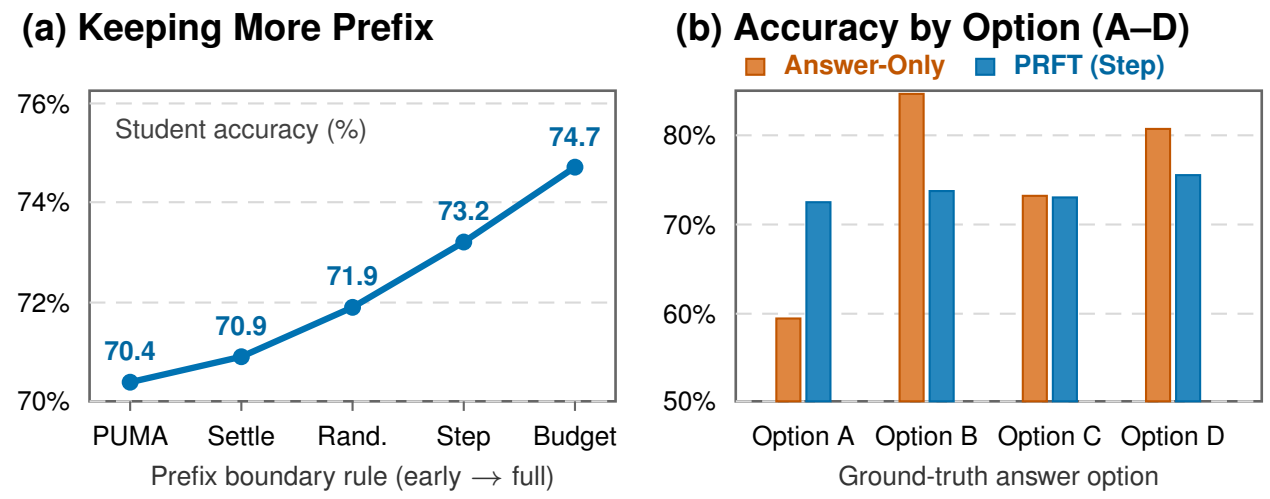
\begin{figure}[t]
  \centering\footnotesize
  \resizebox{\linewidth}{!}{%
  \begin{tikzpicture}[
      x=1cm, y=1.06cm, font=\sffamily\scriptsize,
      lbl/.style={font=\sffamily\scriptsize, text=black!85}
    ]
    % -------- Panel (a): Keeping More Prefix Improves Accuracy --------
    \node[font=\sffamily\scriptsize\bfseries, anchor=west] at (0.0, 2.78) {(a) Keeping More Prefix};
    \draw[black!60, line width=0.5pt] (0.55, 0.45) rectangle (4.05, 2.40);
    \foreach \y/\ylbl in {0.45/70, 1.07/72, 1.70/74, 2.32/76} {
      \draw[black!15, densely dashed] (0.55, \y) -- (4.05, \y);
      \node[font=\sffamily\tiny, anchor=east] at (0.52, \y) {\ylbl\%};
    }
    \foreach \x/\xlbl in {0.82/PUMA, 1.56/Settle, 2.30/Rand., 3.04/Step, 3.78/Budget} {
      \draw[black!60] (\x, 0.45) -- (\x, 0.39);
      \node[font=\sffamily\tiny, anchor=north] at (\x, 0.39) {\xlbl};
    }
    \node[font=\sffamily\tiny, text=black!80, anchor=north] at (2.30, 0.14) {Prefix boundary rule (early $\to$ full)};
    \draw[qlbblue, line width=1.2pt] (0.82, 0.57) -- (1.56, 0.73) -- (2.30, 1.04) -- (3.04, 1.45) -- (3.78, 1.92);
    \foreach \x/\y/\val in {0.82/0.57/70.4, 1.56/0.73/70.9, 2.30/1.04/71.9, 3.04/1.45/73.2, 3.78/1.92/74.7} {
      \fill[qlbblue] (\x, \y) circle (1.6pt);
      \node[font=\sffamily\tiny\bfseries, text=qlbblue!90!black, anchor=south] at (\x, {\y+0.04}) {\val};
    }
    \node[font=\sffamily\tiny, text=black!75, anchor=north west] at (0.62, 2.32) {Student accuracy (\%)};

    % -------- Panel (b): Per-Option Accuracy Across A-D (Main) --------
    \node[font=\sffamily\scriptsize\bfseries, anchor=west] at (4.35, 2.78) {(b) Accuracy by Option (A--D)};
    \draw[black!60, line width=0.5pt] (4.85, 0.45) rectangle (8.35, 2.40);
    \foreach \y/\ylbl in {0.45/50, 1.00/60, 1.56/70, 2.12/80} {
      \draw[black!15, densely dashed] (4.85, \y) -- (8.35, \y);
      \node[font=\sffamily\tiny, anchor=east] at (4.82, \y) {\ylbl\%};
    }
    % Exact verified per-option accuracies (Main 1,200 x 4):
    % Answer-Only (gamma_ans): A=59.3 (0.97), B=84.7 (2.38), C=73.2 (1.74), D=80.7 (2.16)
    % PRFT Step Cut (gamma):   A=72.4 (1.70), B=73.8 (1.77), C=73.1 (1.73), D=75.6 (1.87)
    \foreach \idx/\opt/\ya/\yp in {0/A/0.97/1.70, 1/B/2.38/1.77, 2/C/1.74/1.73, 3/D/2.16/1.87} {
      \pgfmathsetmacro{\xbase}{5.30 + \idx*0.82}
      \fill[qlbverm!75, draw=qlbverm!90!black] ({\xbase-0.18}, 0.45) rectangle ({\xbase-0.02}, \ya);
      \fill[qlbblue!85, draw=qlbblue!95!black] ({\xbase+0.02}, 0.45) rectangle ({\xbase+0.18}, \yp);
      \node[font=\sffamily\tiny, anchor=north] at (\xbase, 0.39) {Option \opt};
    }
    \node[font=\sffamily\tiny, text=black!80, anchor=north] at (6.60, 0.14) {Ground-truth answer option};
    \fill[qlbverm!75, draw=qlbverm!90!black] (4.92, 2.50) rectangle (5.04, 2.60);
    \node[font=\sffamily\tiny\bfseries, text=qlbverm!90!black, anchor=west] at (5.07, 2.55) {Answer-Only};
    \fill[qlbblue!85, draw=qlbblue!95!black] (6.45, 2.50) rectangle (6.57, 2.60);
    \node[font=\sffamily\tiny\bfseries, text=qlbblue!90!black, anchor=west] at (6.60, 2.55) {\method{} (Step)};
  \end{tikzpicture}%
  }
  \vspace{-2.0mm}
  \caption{\textbf{Effect of kept prefix length and prefix supervision.}
    \emph{(a)} Keeping a more complete reasoning prefix steadily improves student accuracy across boundary rules ($70.4\% \to 74.7\%$).
    \emph{(b)} Training on answers without the prefix (\texttt{Answer-Only}) induces severe imbalance across gold options A--D ($25.4\%$ spread), whereas \method{} (Step Cut) stays balanced ($3.2\%$ spread).}
  \label{fig:ablation-bias}
  \vspace{-2.0mm}
\end{figure}

\begin{table}[t]
  \centering\footnotesize
  \setlength{\tabcolsep}{2.6pt}
  \renewcommand{\arraystretch}{0.92}
  \resizebox{\linewidth}{!}{%
  \begin{tabular}{@{}lcccccccc@{}}
    \toprule
    & \multicolumn{7}{c}{Test-Time Thinking Budget (Tokens)} & \\
    \cmidrule(lr){2-8}
    Method (\textsc{Qwen3.5-4B}) & $512$ & $1024$ & $1536$ & $2048$ & $2560$ & $3072$ & Cap & Think Tok \\
    \midrule
    Untrained Base & -- & 66.7 & 69.0 & 69.9 & 70.7 & 71.3 & 73.2 & 5,649 \\
    Standard RFT (SFT) & 64.8 & 68.7 & 70.8 & 72.1 & 72.5 & 72.8 & 73.9 & 4,441 \\
    \textsc{GRPO} (RL, 2 seeds) & 65.9 & 69.3 & 70.3 & 70.8 & 71.3 & 71.3 & 71.2 & 2,564 \\
    \textsc{DAPO} (RL) & -- & 68.9 & 70.4 & 71.3 & 71.8 & 71.5 & 72.0 & 2,084 \\
    Prefix Credit Only & 65.5 & 69.4 & 70.5 & 71.3 & 71.5 & 71.4 & 71.6 & 2,131 \\
    \midrule
    \method{} (Full Pool) & -- & 71.4 & 72.5 & 73.0 & 73.0 & 73.1 & 73.2 & \textbf{1,849} \\
    \method{} (Step Cut) & \textbf{68.8} & \textbf{72.5} & \textbf{73.5} & \textbf{73.7} & \textbf{73.8} & 73.9 & 73.8 & 2,188 \\
    \textbf{\method{} (Budget Cut)} & -- & \textbf{72.5} & 73.4 & \textbf{73.7} & \textbf{73.8} & \textbf{74.3} & \textbf{74.6} & 2,257 \\
    \bottomrule
  \end{tabular}%
  }
  \vspace{-1.8mm}
  \caption{\textbf{Student accuracy (\%) across test-time thinking budgets on Main ($1{,}200$ items).}
    `--' indicates budgets not executed for that checkpoint. \method{} outperforms Standard RFT, online RL (\textsc{GRPO}, \textsc{DAPO}), and Prefix Credit Only at every measured budget, reaching near-ceiling accuracy by $1{,}024$--$1{,}536$ tokens.}
  \label{tab:matched-budget}
  \vspace{-1.5mm}
\end{table}

\begin{table}[t]
  \centering\footnotesize
  \setlength{\tabcolsep}{2.8pt}
  \renewcommand{\arraystretch}{0.92}
  \resizebox{\linewidth}{!}{%
  \begin{tabular}{@{}lcccc@{}}
    \toprule
    \textbf{(a) Teacher Sampling (\textsc{27B}, $9{,}702$ q)} & Solved Q. & Pass Rate (\%) & Sampling Cost & Pass Gain \\
    \midrule
    Standard RFT ($n{=}1$) & 4,583 & 47.2 & $1.00\times$ & Baseline \\
    Standard RFT ($n{=}2$ resampling) & 5,203 & 53.6 & $2.00\times$ & $+6.4\%$ \\
    \method{} ($n{=}1$, prefix probe) & 7,620 & 78.5 & $1.00\times$ & $+31.3\%$ \\
    \textbf{\method{} ($n{=}2$, Budget Cut)} & \textbf{8,337} & \textbf{85.9} & \textbf{$1.00\times$ / rollout} & \textbf{+38.7\%} \\
    \midrule
    \textbf{(b) Teacher Token Limit ($2{,}000$ q)} & Mean Tok & RFT Pass (\%) & \textbf{\method{} Pass (\%)} & Rescued \\
    \midrule
    2,048 & 1,719 & 41.6 & 49.9 & +166 \\
    3,072 (Default) & 2,306 & 52.0 & \textbf{77.8} & \textbf{+516} \\
    4,096 & 2,798 & 58.2 & 78.7 & +409 \\
    8,192 & 4,237 & 72.0 & 81.0 & +181 \\
    12,000 ($2.2\times$ compute) & 5,071 & 79.0 & \textbf{83.5} & \textbf{+90} \\
    \midrule
    \textbf{(c) Teacher Model Family} & Pool / Style & RFT Pass (\%) & \textbf{\method{} Pass (\%)} & Rescued \\
    \midrule
    \textsc{Qwen3.5-27B} & $11.7\text{k}$ / Long think & 56.9 & \textbf{80.0} & \textbf{+2,740} \\
    \textsc{Qwen3.5-9B} & $11.7\text{k}$ / Std.\ think & 51.7 & \textbf{72.5} & \textbf{+2,478} \\
    \textsc{Qwen3.5-35B-A3B} & $9.4\text{k}$ ($n{=}1$) / Ext.\ think & 54.9 & -- & -- \\
    \textsc{InternVL2-26B} & $11.7\text{k}$ / Concise CoT & 55.6 & \textbf{55.7} & +9 \\
    \bottomrule
  \end{tabular}%
  }
  \vspace{-1.5mm}
  \caption{\textbf{Ablation of teacher sampling efficiency, token limits, and teacher model families.}
    \emph{(a)} Single-step prefix probing allows $n{=}1$ \method{} sampling ($78.5\%$) to beat $n{=}2$ RFT resampling ($53.6\%$) at half the sampling cost ($85.9\%$ ungated / $84.2\%$ at $\tau{=}0.6$ for $n{=}2$).
    \emph{(b)} \method{} at a $3{,}072$-token limit matches the pass rate of Standard RFT at a $12{,}000$-token limit.
    \emph{(c)} Prefix Reward rescues over $2{,}400$ unfinished traces on thinking teachers while leaving concise \textsc{InternVL2-26B} traces unchanged (on the matched $9{,}366$-question $n{=}1$ check, \textsc{35B-A3B} endpoint pass rate drops to $54.9\%$ vs.\ $62.9\%$ for \textsc{27B} due to $97.5\%$ unclosed discards).}
  \label{tab:coverage}
  \label{tab:three-teacher}
  \label{tab:capsweep}
  \vspace{-1.5mm}
\end{table}

\begin{figure*}[t]
  \centering\footnotesize
  \setlength{\tabcolsep}{3.5pt}
  \renewcommand{\arraystretch}{1.00}
  \resizebox{\textwidth}{!}{%
  \begin{tabular}{@{}p{0.236\textwidth}p{0.236\textwidth}p{0.236\textwidth}p{0.236\textwidth}@{}}
    \toprule
    \textbf{(a) Teacher Rescue via Prefix Reward} &
    \textbf{(b) Student $15.4\times$ Token Compression} &
    \textbf{(c) Unseen Domain Generalization} &
    \textbf{(d) Cue-Conditional Tail Disagreement} \\
    \scriptsize Physics Scatter Plot (Main Pool) &
    \scriptsize Quantum Spin-Chain Plot (Main Test) &
    \scriptsize Fluid-Dynamics Plot (Unseen Reserve) &
    \scriptsize Combinatorics Graph (9B Census) \\
    \midrule
    \multicolumn{1}{c}{\includegraphics[height=2.25cm,keepaspectratio]{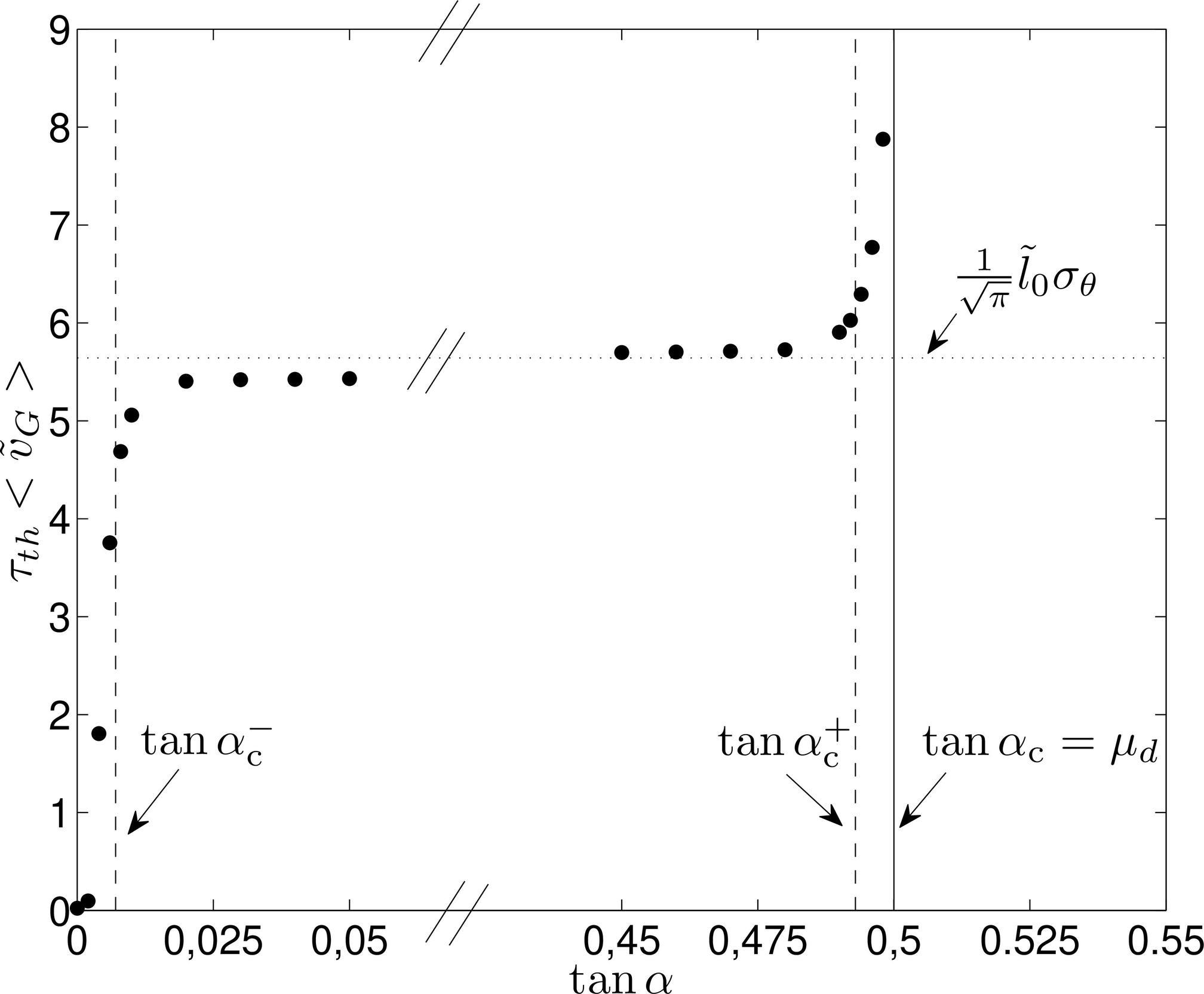}} &
    \multicolumn{1}{c}{\includegraphics[height=2.25cm,keepaspectratio]{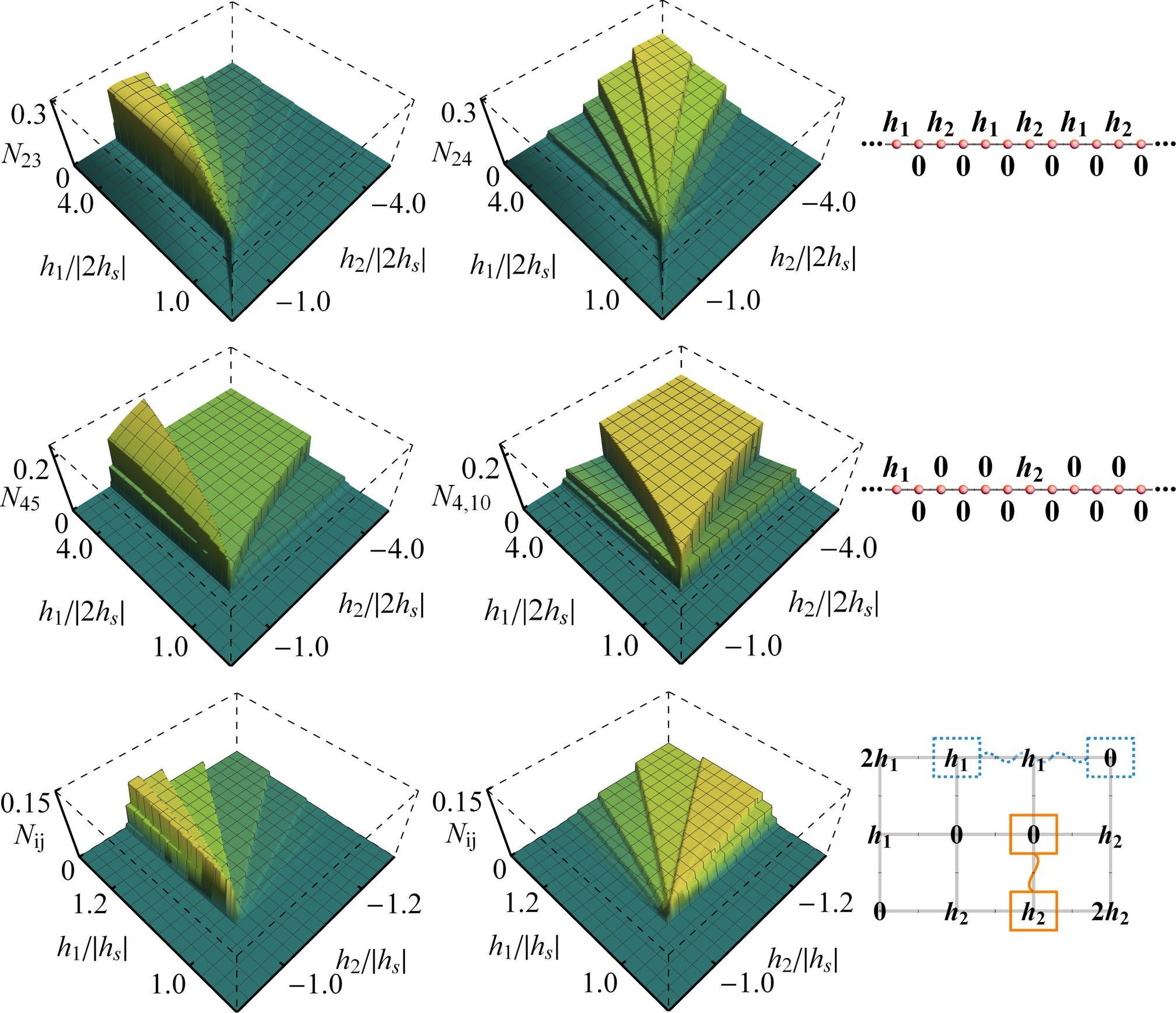}} &
    \multicolumn{1}{c}{\includegraphics[height=2.25cm,keepaspectratio]{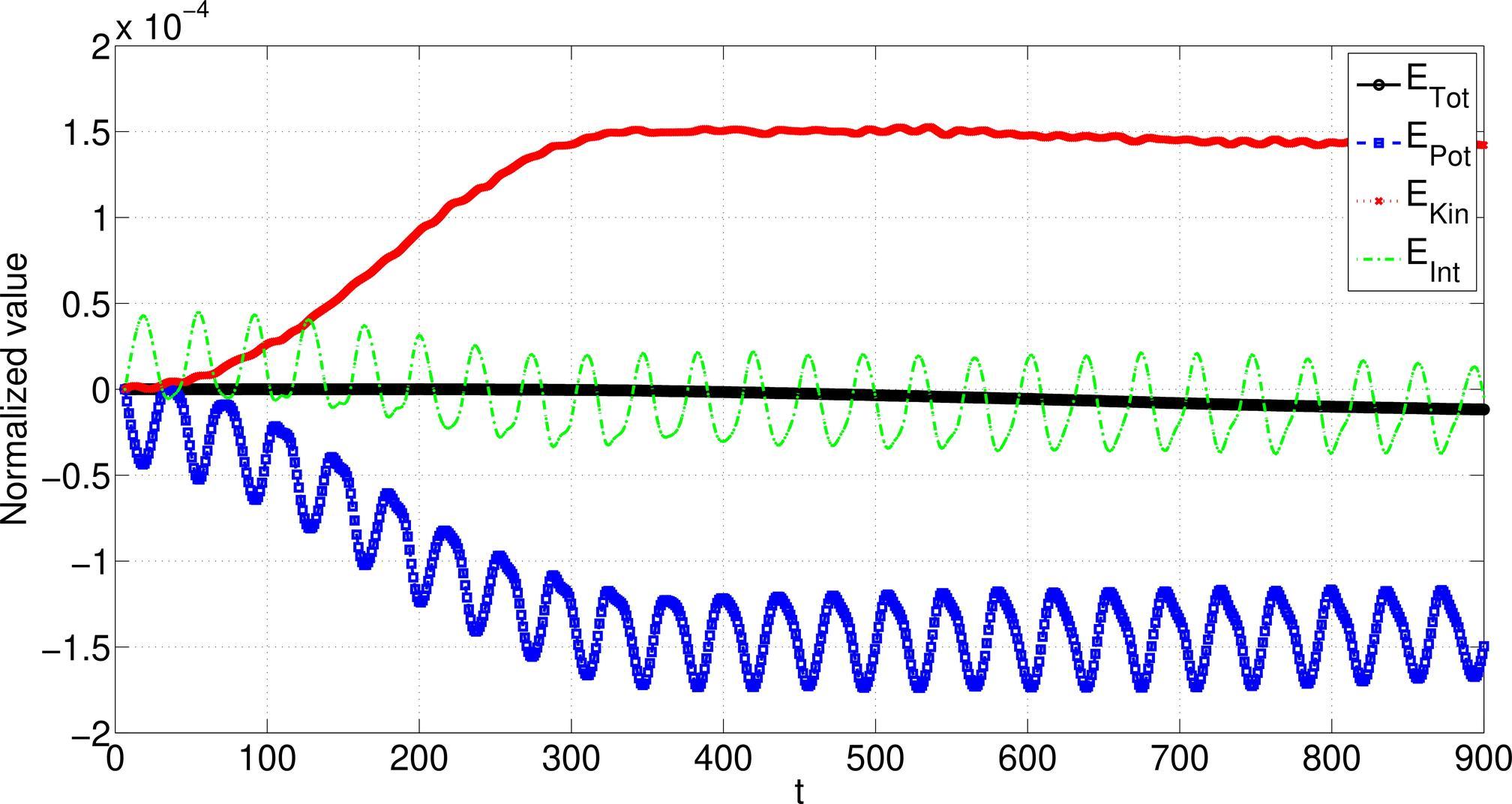}} &
    \multicolumn{1}{c}{\includegraphics[height=2.25cm,keepaspectratio]{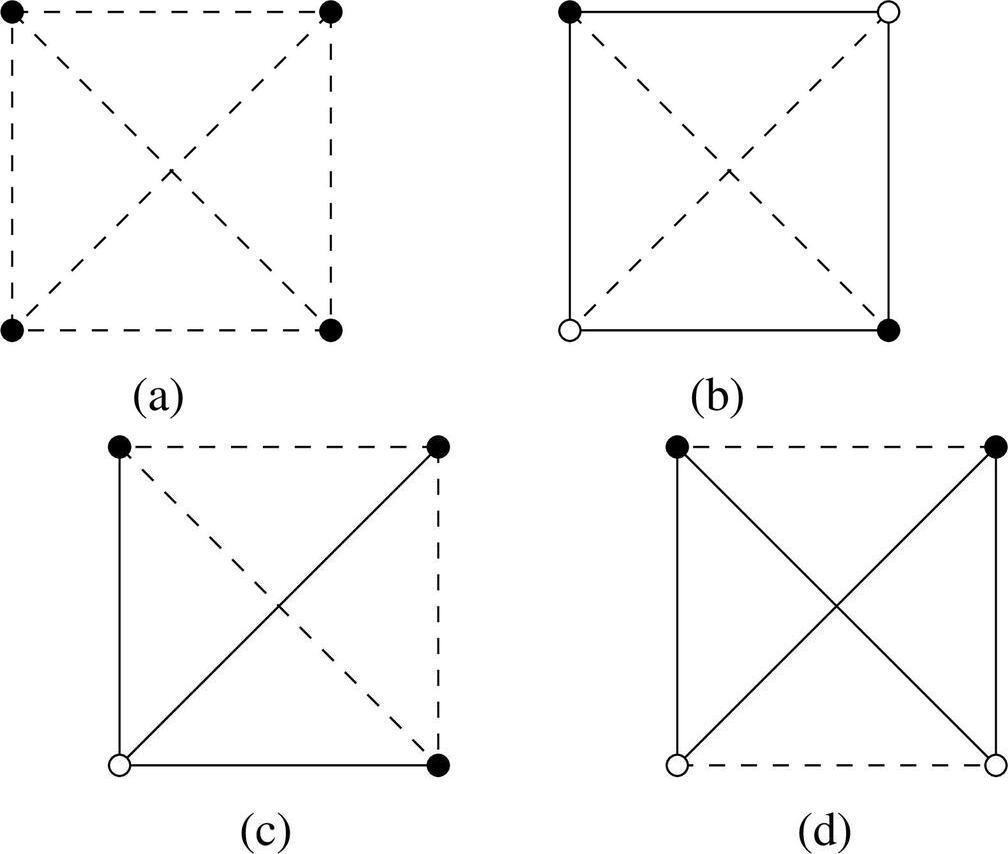}} \\
    \midrule
    \scriptsize \textbf{Q:} What condition defines the critical angle $\alpha_{\mathrm{c}}^+$? (\textbf{Gold: B}, $\mu_{\mathrm{d}}-\sigma_\mu/\sqrt{2}$) &
    \scriptsize \textbf{Q:} At which plateau does pair negativity $N_{45}$ peak? (\textbf{Gold: D}, $M{=}1$) &
    \scriptsize \textbf{Q:} Which energy shows a sustained increment in simulation? (\textbf{Gold: A}, Kinetic) &
    \scriptsize \textbf{Q:} How many crossings occur in figure (c) with 3 black, 1 white point? (\textbf{Gold: B}, $1$) \\
    \midrule
    \scriptsize \textcolor{qlbverm}{\textbf{Standard RFT (Discarded):}} Transcribes plot axes, $\tan\alpha_{\mathrm{c}}^+ {<} 0.5$, and options by tok $448$, then loops on integrals for $2{,}880$ tok until hitting the $3{,}328$ limit. &
    \scriptsize \textcolor{qlbverm}{\textbf{Standard RFT (Failed):}} Repeats 12-spin index checks and hits the $12{,}000$-tok limit ($11{,}792$ mean tok), predicting wrong \texttt{B} under $B{=}512$. &
    \scriptsize \textcolor{qlbverm}{\textbf{Standard RFT (Failed):}} Notes red $E_{\mathrm{Kin}}$ curve rises, yet loops to the $12{,}000$-tok limit and mis-selects \texttt{C} (Internal) under $B{=}512$. &
    \scriptsize \textcolor{qlbverm}{\textbf{Capped Endpoint Readout:}} Early prefix yields \texttt{B}, but $5{,}500{+}$ tok of tail checking end in length truncation where forced readout at tok $8{,}192$ outputs \texttt{C}. \\
    \scriptsize \textcolor{qlbgreen!80!black}{\textbf{\method{} (Ours):}} Probes cached prefix at tok $448$ ($\hat{y}_k{=}\text{\texttt{B}}, p_k{=}0.83$), masks the $2{,}880$-tok loop, and yields a clean \textbf{456-tok ($7.3\times$ shorter)} target. &
    \scriptsize \textcolor{qlbgreen!80!black}{\textbf{\method{} (Ours):}} Locates panel $N_{45}$ and peak $M{=}1$ within $512$ tok, predicting \texttt{D} and stopping cleanly in \textbf{779 tok ($15.4\times$ shorter)}. &
    \scriptsize \textcolor{qlbgreen!80!black}{\textbf{\method{} (Ours):}} Aligns early curve reading with readout, predicting \texttt{A} (Kinetic) and stopping cleanly in \textbf{1,697 tok ($7.3\times$ shorter)}. &
    \scriptsize \textcolor{qlbgreen!80!black}{\textbf{\method{} (Ours):}} Probes prefix at tok $2{,}628$ and locks in \texttt{B} ($1$ crossing) in \textbf{2,636 tok ($3.1\times$ shorter)} before length-capped drift. \\
    \bottomrule
  \end{tabular}%
  }
  \vspace{-1.5mm}
  \caption{\textbf{Qualitative comparison across four visual reasoning scenarios.}
    \emph{(a)} Prefix Reward rescues an unfinished teacher rollout while Prefix Credit Assignment removes the looping tail.
    \emph{(b, c)} \method{} compresses student generation by $15.4\times$ and $7.3\times$ while solving main and unseen Reserve figures within a $512$-token budget.
    \emph{(d)} Prolonged tail verification hits the $8{,}192$-token limit where cue-conditional endpoint readout disagrees with the correct prefix readout (\texttt{B} vs.\ \texttt{C}), whereas \method{} locks in the prefix target before the cap.}
  \label{fig:case-studies}
  \vspace{-1.5mm}
\end{figure*}

\paragraph{Both Prefix Reward and Prefix Credit Assignment are essential.}
\Cref{tab:ablation}a isolates the effect of each design in \method{}.
Applying Prefix Credit Assignment alone on naturally finished teacher traces (\texttt{Prefix Credit Only}) removes tail-loop supervision and halves student thinking tokens ($4{,}441 \to 2{,}131$), lifting SFT accuracy by \textbf{+3.3\%} to match online RL.
This shows that standard SFT trails online RL largely because uniform full-trace supervision teaches the student to repeat tail loops.
Adding Prefix Reward further lifts accuracy to \textbf{73.7\%} at matched rows/steps (\textbf{74.7\%} with Budget Cut, and pairs cleanly with optional Visual-Gain Weighting at $74.2\%$ or Vision LoRA at $74.5\%$), outperforming unfiltered SFT ($71.9\%$) and matched training controls (\cref{app:impl}).
Prefix Credit Assignment and Prefix Reward are thus complementary: the former teaches concise termination, while the latter supplies supervision on hard visual figures.

\paragraph{Keeping the full prefix improves accuracy and avoids option bias.}
\Cref{tab:ablation}b and \cref{fig:ablation-bias}a examine where to truncate rescued teacher rollouts.
Truncating at the earliest correct or stable probe (\textsc{PUMA}, \textsc{Settle}) lowers student accuracy ($70.4\%$--$70.9\%$) by discarding intermediate steps that link visual observations to the final answer.
Even after the teacher's prediction stabilizes, retaining reasoning steps up to the prefix budget steadily improves student accuracy from $70.4\%$ to \textbf{74.7\%}.
At the opposite extreme, removing the reasoning prefix altogether (\texttt{Answer-Only SFT}) induces severe option-letter bias, creating a $25.4\%$ accuracy gap across gold options A--D compared with $3.2\%$ for \method{} (Step Cut; \cref{fig:ablation-bias}b).
Preserving a complete deduction prefix is therefore crucial to teach balanced visual reasoning.

\paragraph{\method{} remains accurate under tight token budgets.}
\Cref{tab:matched-budget} and \cref{fig:exp_summary}a evaluate students across test-time thinking budgets by forcing an answer readout whenever generation reaches the budget limit.
\method{} outperforms Standard RFT, online RL, and Prefix Credit Only across every measured budget, reaching near-ceiling accuracy by $1{,}024$--$1{,}536$ tokens.
Notably, under the tightest $512$-token budget, \method{} (Step Cut) leads Standard RFT by \textbf{+4.0\%} ($68.8\%$ vs.\ $64.8\%$; 3-seed mean $+3.3\%$, \cref{tab:multiseed-cap}a) and \textsc{GRPO} by \textbf{+2.9\%}, with $99\%$ of the gain over Standard RFT occurring on items where both models are probed at exact token $512$.
This indicates that \method{} not only teaches cleaner stopping, but also makes early reasoning steps more informative.

\paragraph{Prefix probing beats resampling and longer token limits.}
A natural question is whether endpoint filtering can be fixed simply by drawing more teacher rollouts or increasing the teacher's token limit (\cref{tab:coverage}).
In \cref{tab:coverage}a, doubling the number of teacher rollouts under Standard RFT yields only $+6.4\%$ ($47.2\% \to 53.6\%$) because visual ambiguities trap the teacher in repetitive tail loops across repeated trials.
By contrast, probing the cached prefix in a single step raises $n{=}1$ pass rate to $78.5\%$ and $n{=}2$ Budget-Cut pass rate to \textbf{85.9\%} (\cref{fig:exp_summary}c).
Similarly, \cref{tab:capsweep}b shows that Standard RFT requires a $4\times$ longer generation limit ($12{,}000$ tokens, $2.2\times$ mean decode tokens) just to match the pass rate that \method{} achieves at $3{,}072$ tokens.
Across thinking teachers (\cref{tab:three-teacher}c), Prefix Reward rescues over $2{,}400$ unfinished traces per pool while leaving concise \textsc{InternVL2-26B} traces essentially unchanged ($+9$ format-only rescues).

\subsection{Visualization and Mechanism Analysis}
\label{sec:exp:mechanism}
\label{sec:exp:vis}

\begin{figure}[t]
  \centering\footnotesize
  \resizebox{\linewidth}{!}{%
  \begin{tikzpicture}[
      x=1cm, y=1.52cm, font=\sffamily\scriptsize,
      lbl/.style={font=\sffamily\scriptsize, text=black!85}
    ]
    % -------- Panel (a): Image Reliance Across Reasoning --------
    \node[font=\sffamily\scriptsize\bfseries, anchor=west] at (0.0, 2.78) {(a) Image Reliance in Reasoning};
    \draw[black!60, line width=0.5pt] (0.55, 0.45) rectangle (4.05, 2.40);
    \foreach \y/\ylbl in {0.45/0, 0.94/25, 1.43/50, 1.91/75, 2.40/100} {
      \draw[black!15, densely dashed] (0.55, \y) -- (4.05, \y);
      \node[font=\sffamily\tiny, anchor=east] at (0.52, \y) {\ylbl\%};
    }
    \foreach \x/\xlbl in {0.75/10\%, 1.37/20\%, 1.99/40\%, 2.61/60\%, 3.23/80\%, 3.85/100\%} {
      \draw[black!60] (\x, 0.45) -- (\x, 0.39);
      \node[font=\sffamily\tiny, anchor=north] at (\x, 0.39) {\xlbl};
    }
    \node[font=\sffamily\tiny, text=black!80, anchor=north] at (2.30, 0.14) {Reasoning progress (prefix $\to$ tail)};
    % Exact normalized positional Delta_t across deciles 1,2,4,6,8,10: 100.0, 61.5, 38.1, 29.9, 28.4, 12.4
    \draw[qlbblue, line width=1.2pt] (0.75, 2.40) -- (1.37, 1.65) -- (1.99, 1.19) -- (2.61, 1.03) -- (3.23, 1.00) -- (3.85, 0.69);
    \foreach \x/\y in {0.75/2.40, 1.37/1.65, 1.99/1.19, 2.61/1.03, 3.23/1.00, 3.85/0.69} {
      \fill[qlbblue] (\x, \y) circle (1.5pt);
    }
    % Exact visual sufficiency drop w/o image (n=500 quantiles: 30.2, 4.8, 2.2, 0.9, 0.4, 0.7 -> y = 0.45 + v/100*1.95)
    \draw[qlbverm, densely dashed, line width=1.1pt] (0.75, 1.04) -- (1.37, 0.54) -- (1.99, 0.49) -- (2.61, 0.47) -- (3.23, 0.46) -- (3.85, 0.46);
    \foreach \x/\y in {0.75/1.04, 1.37/0.54, 1.99/0.49, 2.61/0.47, 3.23/0.46, 3.85/0.46} {
      \fill[qlbverm] (\x, \y) circle (1.4pt);
    }
    \draw[qlbblue, line width=1.2pt] (1.35, 2.22) -- (1.65, 2.22);
    \fill[qlbblue] (1.50, 2.22) circle (1.5pt);
    \node[font=\sffamily\tiny\bfseries, text=qlbblue!90!black, anchor=west] at (1.70, 2.22) {Image Reliance};
    \draw[qlbverm, densely dashed, line width=1.1pt] (1.35, 1.98) -- (1.65, 1.98);
    \fill[qlbverm] (1.50, 1.98) circle (1.4pt);
    \node[font=\sffamily\tiny\bfseries, text=qlbverm!90!black, anchor=west] at (1.70, 1.98) {Drop w/o Image};

    % -------- Panel (b): Non-Termination With vs. Without Figure (Paired TOK stack) --------
    \node[font=\sffamily\scriptsize\bfseries, anchor=west] at (4.35, 2.78) {(b) Non-Termination Rate (\%)};
    \draw[black!60, line width=0.5pt] (4.85, 0.45) rectangle (8.35, 2.40);
    \foreach \y/\ylbl in {0.45/0, 1.07/5, 1.70/10, 2.32/15} {
      \draw[black!15, densely dashed] (4.85, \y) -- (8.35, \y);
      \node[font=\sffamily\tiny, anchor=east] at (4.82, \y) {\ylbl\%};
    }
    % Exact paired TOK values: Std RFT 13.6 vs 0.8 (y=2.15, 0.55); Prefix Credit 1.0 vs 0.6 (y=0.57, 0.52); PRFT Step Cut 1.2 vs 0.4 (y=0.60, 0.50)
    \foreach \idx/\mlbl/\yw/\ywo/\vw/\vwo/\goff in {
      0/Std.\ RFT/2.15/0.55/13.6/0.8/0,
      1/Prefix Credit/0.57/0.52/1.0/0.6/0.11,
      2/\method{} (Step)/0.60/0.50/1.2/0.4/0.09
    } {
      \pgfmathsetmacro{\xbase}{5.45 + \idx*1.15}
      \fill[qlbverm!75, draw=qlbverm!90!black] ({\xbase-0.22}, 0.45) rectangle ({\xbase-0.02}, \yw);
      \fill[qlbblue!80, draw=qlbblue!95!black] ({\xbase+0.02}, 0.45) rectangle ({\xbase+0.22}, \ywo);
      \node[font=\sffamily\tiny\bfseries, text=qlbverm!90!black, anchor=south] at ({\xbase-0.12}, \yw) {\vw};
      \node[font=\sffamily\tiny, text=qlbblue!90!black, anchor=south] at ({\xbase+0.12}, {\ywo+\goff}) {\vwo};
      \node[font=\sffamily\tiny, anchor=north] at (\xbase, 0.39) {\mlbl};
    }
    \node[font=\sffamily\tiny, text=black!80, anchor=north] at (6.60, 0.14) {Student training method};
    \fill[qlbverm!75, draw=qlbverm!90!black] (4.92, 2.50) rectangle (5.04, 2.60);
    \node[font=\sffamily\tiny\bfseries, text=qlbverm!90!black, anchor=west] at (5.07, 2.55) {With Figure};
    \fill[qlbblue!80, draw=qlbblue!95!black] (6.45, 2.50) rectangle (6.57, 2.60);
    \node[font=\sffamily\tiny\bfseries, text=qlbblue!90!black, anchor=west] at (6.60, 2.55) {Without Figure};
  \end{tikzpicture}%
  }
  \vspace{-2.0mm}
  \caption{\textbf{Early image reliance and figure-induced tail looping.}
    \emph{(a)} Teacher per-token log-probability gain from the image ($\Delta_t / \Delta_1$, $11.8\text{M}$ tokens) drops by $8\times$ from the first decile ($100\%$) to the tail ($12.4\%$); masking the image during readout drops accuracy sharply at $0$ tokens ($25.7\%$ on $n{=}2{,}000$; $30.2\%$ on $n{=}500$) but by $<1\%$ once the prefix is written ($0.5\%$ at last boundary).
    \emph{(b)} On paired \texttt{TOK}-stack evaluations, removing the figure cuts Standard RFT non-termination by $18\times$ ($13.6\% \to 0.8\%$), confirming that non-terminating tails are figure-induced and largely eliminated by \method{} ($1.2\%$ Step Cut; $0.04\%$ Budget Cut).}
  \label{fig:vis-decay}
  \vspace{-3mm}
\end{figure}
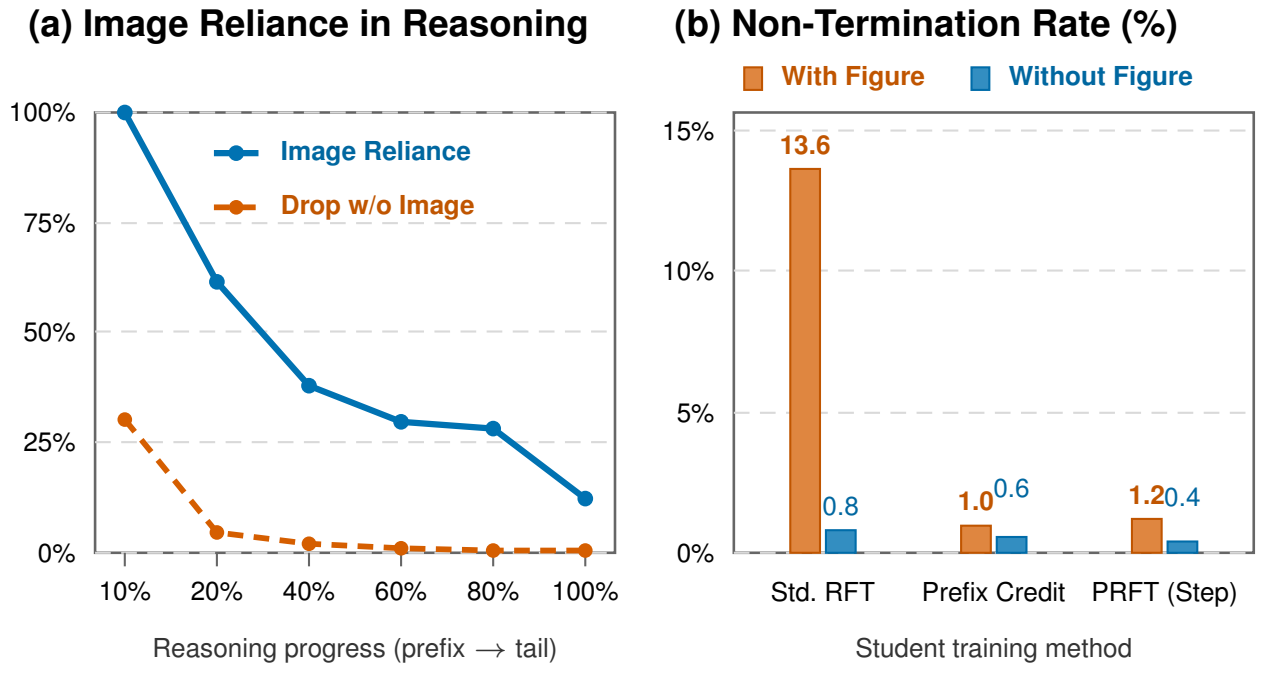

\paragraph{Visual deduction finishes early in the prefix.}
Qualitative examples in \cref{fig:case-studies} illustrate our core observation: on complex scientific plots, the teacher extracts relevant visual cues early in the prefix, whereas the remaining generation consists of repetitive self-checking that hits the token limit (\cref{fig:case-studies}a--c) or leads to cue-conditional readout disagreement at the cap (\cref{fig:case-studies}d).
This pattern is confirmed quantitatively in \cref{fig:vis-decay}a: across $11.8\text{M}$ teacher tokens, relative image reliance ($\Delta_t / \Delta_1$) drops by $8\times$ toward the tail ($100.0\% \to 12.4\%$), and masking the figure during answer readout after the prefix has been written changes accuracy by only $0.5\%$ (compared with $25.7\%$ at zero thinking tokens).
Furthermore, removing the figure at test time cuts Standard RFT non-termination by $18\times$ ($13.6\% \to 0.8\%$, \cref{fig:vis-decay}b), confirming that non-terminating tail loops are tied to visual inputs and cleanly suppressed by prefix supervision.

\begin{table}[t]
  \centering\footnotesize
  \setlength{\tabcolsep}{3.6pt}
  \renewcommand{\arraystretch}{1.05}
  \resizebox{\linewidth}{!}{%
  \begin{tabular}{@{}lcccc@{}}
    \toprule
    Probing Rule & Probes / Trace & Pass Rate (\%) & H100 Time (s) & Relative Time \\
    \midrule
    Endpoint Filter & 0.00 & 53.6 & -- & -- \\
    \textsc{PUMA} (token thirds)~\cite{puma2026} & 3.00 & 89.0 & 66.1 & $1.73\times$ \\
    \textsc{Settle} (all bound.\ + $T$)~\cite{settle2026} & 10.54 & 84.3 & 111.5 & $2.90\times$ \\
    \textbf{\method{} (Ours, single probe)} & \textbf{1.00} & \textbf{85.9} & \textbf{38.4} & \textbf{1.00$\times$} \\
    \bottomrule
  \end{tabular}%
  }
  \vspace{-1.5mm}
  \caption{\textbf{Probing efficiency across early-exit position rules.}
    Pass rates on the $9{,}702$-question \textsc{Qwen3.5-27B} pool (ungated) and offline H100 probing wall times on the $64$-trace 4-GPU rotation panel. A single prefix probe captures $85.9\%$ pass rate ($93.6\%$ of the $12.6$-probe union ceiling) while running $1.73\times$--$2.90\times$ faster than multi-probe rules.}
  \label{tab:mechanism}
  \label{tab:offline-config}
  \vspace{-3mm}
\end{table}

\paragraph{One-step probing is fast and concentrates gradients on the prefix.}
Because visual information is already transcribed into text by the prefix boundary, \method{} requires only a single cached probe per unfinished rollout (\cref{tab:mechanism}), achieving a higher pass rate than \textsc{Settle}'s $10.54$-probe rule ($85.9\%$ vs.\ $84.3\%$) while running $2.90\times$ faster on H100 GPUs ($1.73\times$ faster than \textsc{PUMA}'s 3-probe rule).
Furthermore, local gradient diagnostics at trained checkpoints (\cref{app:latency}) show that the standardized stop-and-answer suffix on cut targets reaches near-zero cross-entropy ($\sim 6\times 10^{-5}$), leaving roughly $99\%$ of the local LoRA gradient norm on the reasoning prefix.

\section{Conclusion}
\label{sec:conclusion}

We presented \textbf{Prefix Rejection Fine-Tuning (\method{})}, a simple and general prefix-supervised SFT framework for multimodal reasoning distillation.
Standard endpoint-based SFT and RFT break down on complex scientific figures because visual deduction finishes early in the reasoning prefix whereas fine-grained visual ambiguities trigger repetitive tail checking, causing endpoint discard on unfinished rollouts and tail imitation on surviving full traces.
Coupling a cached single-step \textbf{Prefix Reward} with \textbf{Prefix Credit Assignment} evaluates correctness directly from the prefix, removes repetitive tail loops, and appends a clean stop-and-answer transition, training 4B and 9B student models to surpass tested SFT variants and online RL baselines under natural decoding while cutting thinking tokens by roughly half.

\paragraph{Limitations and future work.}
First, our single-step prefix readout targets closed-form and multiple-choice training signals; extending Prefix Reward to open-ended derivations requires multi-token prefix readouts with semantic answer matching (as reflected by \method{}'s lower open-ended score on competition-level \textsc{MathVision} relative to \textsc{DAPO}).
Second, under external forced readout at the full $12{,}000$-token cap, Standard RFT narrows or closes the accuracy gap at roughly $2\times$ the token cost, indicating that \method{}'s primary strength lies in internalizing concise self-terminating reasoning and tight-budget accuracy.
Third, when a frozen vision encoder misbinds dense monochrome line styles or overlapping symbols in scientific plots, prefix pruning eliminates looping but requires vision-side adaptation to resolve perceptual ambiguity.

{
  \small
  \bibliographystyle{ieeenat_fullname}
  \bibliography{main}
}

\maketitleappendix
\appendix
\section{Implementation Details and Training Controls}
\label{app:impl}
\label{app:audit}

\begin{algorithm}[t]
  \caption{\method{}: \methodlong{}}
  \label{alg:harvest}
  \footnotesize
  \begin{algorithmic}[1]
    \Require Question pool $\mathcal{Q} = \{(x_i, y_i)\}_{i=1}^{N}$, teacher rollouts $\{\tau_{i,j}\}_{j=1}^{n}$, prefix budget $k$, base student $P_{\theta_0}$
    \Statex \textbf{Part 1: Single-Step Cached Prefix Readout}
    \For{each rollout $\tau_{i,j} = (t_1, \dots, t_L)$ on $(x_i, y_i) \in \mathcal{Q}$}
      \If{$\tau_{i,j}$ finishes naturally within $k$ tokens}
        \State Set $\hat{y}_k \gets$ terminal answer of $\tau_{i,j}$
      \Else
        \State Branch cached key-value states of $\tau_{i,j,\le k}$ and decode $\hat{y}_k = \operatorname{Readout}(x_i, \tau_{i,j,\le k})$ in one step (\cref{eq:ir})
      \EndIf
      \State Compute Prefix Reward $R_{\mathrm{pre}}(\tau_{i,j}, y_i) = \mathbf{1}[\hat{y}_k = y_i]$ (\cref{eq:prefix_reward})
    \EndFor
    \Statex \textbf{Part 2: Prefix Reward Selection \& Prefix Credit SFT}
    \State Initialize supervision set $\mathcal{D} \gets \varnothing$
    \For{each question $(x_i, y_i) \in \mathcal{Q}$}
      \If{$\max_{1 \le j \le n} R_{\mathrm{pre}}(\tau_{i,j}, y_i) = 1$}
        \State Select shortest positive prefix $\tau_{i,j^\star,\le k}$
        \State Set prefix weights $w_t = 1$ ($t \le k$) and tail weights $w_t = 0$ ($t > k$)
        \State Append clean stop-and-answer transition for $y_i$ and add to $\mathcal{D}$
      \EndIf
    \EndFor
    \State Train student $P_\theta$ on $\mathcal{D}$ with $\mathcal{L}_{\mathrm{\method{}}}(\theta)$ (\cref{eq:admit_gate})
  \end{algorithmic}
\end{algorithm}

\begin{table}[t]
  \centering\footnotesize
  \setlength{\tabcolsep}{2.2pt}
  \renewcommand{\arraystretch}{0.96}
  \resizebox{\linewidth}{!}{%
  \begin{tabular}{@{}lcccccc@{}}
    \toprule
    \rowcolor{fam}[\tabcolsep][\tabcolsep]
    \textbf{Method / Training Control (\textsc{4B})} & \textbf{Train Q.} & \textbf{Steps} & \textbf{Cut (\%)} & \textbf{Acc.\ (\%)} & \textbf{Think Tok} & \textbf{Option Bias} \\
    \midrule
    Direct-Answer SFT & 6,656 & 624 & 0.0 & 71.0 & 0 & 6.6\% \\
    Standard RFT~\cite{rftscaling2023} & 6,656 & 624 & 0.0 & 68.0 & 4,441 & 4.8\% \\
    Unfiltered SFT (All Traces, 1 ep.) & 16,944 & 529 & -- & 71.9 & 2,068 & 5.9\% \\
    Answer-Only SFT (w/o Prefix) & 6,656 & 624 & 0.0 & 74.8 & 0 & \textcolor{qlbverm}{\textbf{25.4\%}} \\
    \midrule
    Prefix Credit Only (w/o $R_{\mathrm{pre}}$) & 6,656 & 624 & 11.6 & 71.3 & 2,131 & 2.7\% \\
    Step-Matched Control (Dup.\ Beta) & 9,366 & 876 & 11.6 & 71.8 & 2,238 & 3.5\% \\
    Length-Matched Control & 6,656 & 624 & 47.8 & 73.5 & \textbf{1,677} & 4.6\% \\
    RL (\textsc{DAPO}) $\to$ \method{} (2 seeds) & 6,656 & 624 & 47.8 & 71.2 & 1,765 & \textbf{2.2\%} \\
    \midrule
    \method{} (Full Pool, Step Cut) & 9,366 & 876 & 37.5 & 73.2 & 1,849 & 3.5\% \\
    \rowcolor{ours}[\tabcolsep][\tabcolsep]
    \method{} (Matched-Row, Step Cut) & 6,656 & 624 & 47.8 & 73.7 & 2,188 & 3.2\% \\
    \rowcolor{ours}[\tabcolsep][\tabcolsep]
    \textbf{\method{} (Shared-Pool, Budget Cut)} & 7,621 & 714 & 41.0 & \textbf{74.7} & 2,257 & 6.3\% \\
    \bottomrule
  \end{tabular}%
  }
  \vspace{-1.8mm}
  \caption{\textbf{Training configurations and matched optimization controls on \textsc{Qwen3.5-4B} Main.}
    \textbf{Train Q.}, \textbf{Steps}, and \textbf{Cut (\%)} denote training examples, optimizer updates (3 epochs except Unfiltered SFT at 1 epoch), and the fraction of prefix-cut trajectories.
    At strictly matched question counts and update steps ($6{,}656$ examples, $624$ steps), \method{} (Step Cut) outperforms Standard RFT ($+5.7\%$), Prefix Credit Only ($+2.4\%$), Step-Matched, Length-Matched, and RL-initialized controls.}
  \label{tab:train-configs}
  \vspace{-2.5mm}
\end{table}

\paragraph{Cached prefix readout and boundary rules.}
\Cref{alg:harvest} summarizes the \method{} training pipeline.
Teacher rollouts are sampled from \textsc{Qwen3.5-27B}, \textsc{9B}, \textsc{35B-A3B}~\cite{qwen35}, and \textsc{InternVL2-26B}~\cite{internvl2024,chen2024} in thinking mode ($T=1.0$, $\text{top-}p=0.95$, $\text{top-}k=20$, presence penalty $1.5$), using $n=2$ rollouts per item on the $11{,}702$-question pool ($9{,}702$ up to $3{,}328$ tokens and $2{,}000$ up to $8{,}192$) and $n=8$ on the $1{,}627$-question Hard-Tail Pool.

For any rollout unfinished at the $3{,}072$-token prefix budget, we branch its vLLM key-value cache (\texttt{enable\_prefix\_caching=True}), append \texttt{</think>\textbackslash n\textbackslash n\{"answer": "}, and decode $\hat{y}_k$ in one greedy step over $\{\texttt{A}, \texttt{B}, \texttt{C}, \texttt{D}\}$ with a $0.60$ option-renormalized confidence filter.
Within this budget, \textbf{Step Cut} truncates at the latest paragraph break in $[256, 3{,}072]$, whereas \textbf{Budget Cut} retains all prefix tokens up to $3{,}072$ so traces without early paragraph breaks are also kept.

\paragraph{Student LoRA training and baseline setups.}
All reported students start from the same base checkpoint (\textsc{Qwen3.5-4B} or \textsc{9B}~\cite{qwen35}) and train language-side LoRA~\cite{lora2022} adapters ($r=64$, $\alpha=128$, dropout $0.0$, learning rate $10^{-4}$, cosine schedule, global batch size $32$, $3$ epochs, global token-mean cross-entropy).
By default, Prefix Credit Assignment uses uniform prefix weights ($w_t = 1$ for $t \le k$, $w_t = 0$ for $t > k$) and appends \texttt{</think>\textbackslash n\textbackslash n\{"answer": "}$y$\texttt{"\}}.

In the optional \texttt{Visual-Gain Weighting} variant (\cref{tab:ablation}a), prefix weights follow $w_t \propto 1 + \max(\Delta_t, 0)$ normalized to unit mean per trace, where $\Delta_t = \log P_T(t_t \mid I, q, t_{<t}) - \log P_T(t_t \mid \varnothing, q, t_{<t})$ is the teacher's per-token log-probability gain from figure $I$.
For baselines, \textsc{OPD}~\cite{agarwal2023,minillm2024} applies top-$16$ teacher forward-KL on $9{,}366$ prefixes ($2$ epochs, $584$ steps, $1.47\text{ GiB}$ cached logits with one-hot correction on forced control tokens), \textsc{PUMA}~\cite{puma2026} and \textsc{Settle}~\cite{settle2026} cut at earliest-correct and earliest-stable probes, and \textsc{GRPO}~\cite{shao2024deepseekmath} ($12$ rounds, $12{,}288$ rollouts) and \textsc{DAPO}~\cite{dapo2025} ($6$ rounds, $6{,}144$ rollouts) optimize online group rewards.
Additional cross-family and multi-teacher LoRA fits ($30$ physical checkpoints across \textsc{InternVL2-4B}/\textsc{Phi-3}~\cite{chen2024,phi3technicalreport2024} students and \textsc{Qwen3.5-9B}/\textsc{InternVL2-26B} teachers) completed training, while their full multi-shard evaluation validity gates remain pending and are therefore omitted from downstream accuracy tables.

\paragraph{Matched controls, split boundaries, and integrity audit.}
\Cref{tab:train-configs} lists question counts and update steps across all 4B variants in \cref{tab:main-results,tab:ablation}.
At $6{,}656$ questions and $624$ steps, \texttt{Prefix Credit Only} trims $774$ long endpoint traces ($9.87\text{M}$ target tokens), whereas \texttt{\method{} (Matched-Row, Step Cut)} swaps in $2{,}740$ rescued hard prefixes ($47.8\%$ cut share, $11.81\text{M}$ target tokens) and outperforms Step-Matched ($9{,}366$ duplicated rows, $876$ steps), Length-Matched ($10.42\text{M}$ tokens), and RL-to-\method{} controls.

For evaluation, \texttt{Main} ($1{,}200$ items, $1{,}107$ papers), \texttt{Held-Out} ($1{,}379$ items, $1{,}117$ papers; $1{,}053$ visually dependent), and \texttt{Unseen Reserve} ($1{,}216$ items, $1{,}121$ papers) come from paper-disjoint splits outside the training pool (a blind perceptual-hash scan over $14{,}565$ image paths flagged $1$ cross-paper near-duplicate pair between the training pool and \texttt{Main}).
\texttt{Hard-Tail} ($2{,}306$ items) is the teacher-rejected slice of the $11{,}702$-question candidate pool ($0$ overlap with Standard RFT, Prefix Credit Only, Step Cut, and Full Pool SFT; $12$ items in Shared-Pool cuts, $151$ in \textsc{DAPO} rollouts, and $303$ in \textsc{GRPO} rollouts).
All main scores average $4$ decoding seeds ($12{,}000$-token cap, scoring parsed post-\texttt{</think>} answers) with alphabetical tie-breaking for \textbf{Maj@4}~\cite{selfconsistency2023}.

\section{Multi-Seed Stability and Budget Scaling}
\label{app:matched}
\label{app:frontier2}

\begin{table}[t]
  \centering\footnotesize
  \setlength{\tabcolsep}{2.4pt}
  \renewcommand{\arraystretch}{0.96}
  \resizebox{\linewidth}{!}{%
  \begin{tabular}{@{}lccccc@{}}
    \toprule
    \rowcolor{fam}[\tabcolsep][\tabcolsep]
    \textbf{(a) Three-Seed Stability (\textsc{4B})} & \textbf{Seed 17} & \textbf{Seed 29} & \textbf{Seed 43} & \textbf{3-Seed Mean} & \textbf{Finish (\%)} \\
    \midrule
    \multicolumn{6}{@{}l}{\emph{Main Test ($1{,}200$ items $\times$ $4$ decodes = $4{,}800$ outputs / seed; Natural / $B{=}512$)}} \\
    Standard RFT & 68.0 / 64.8 & 68.2 / 66.5 & 68.5 / 65.7 & 68.2 / 65.7 & 86.7 \\
    Prefix Credit Only & 71.3 / 65.5 & \textbf{72.7} / 67.4 & 73.2 / 67.3 & 72.4 / 66.8 & 98.7 \\
    \rowcolor{ours}[\tabcolsep][\tabcolsep]
    \textbf{\method{} (Step Cut)} & \textbf{73.7 / 68.8} & 72.6 / \textbf{68.9} & \textbf{73.6 / 69.2} & \textbf{73.3 / 68.9} & \textbf{99.5} \\
    \midrule
    \multicolumn{6}{@{}l}{\emph{Unseen Reserve ($1{,}216$ items $\times$ $4$ decodes = $4{,}864$ outputs / seed; Natural / $B{=}512$)}} \\
    Standard RFT & 69.0 / 66.9 & 70.4 / 67.5 & 69.0 / 67.7 & 69.4 / 67.4 & 86.6 \\
    Prefix Credit Only & 72.0 / 68.4 & \textbf{72.9} / 67.8 & 73.0 / 68.4 & 72.6 / 68.2 & 98.5 \\
    \rowcolor{ours}[\tabcolsep][\tabcolsep]
    \textbf{\method{} (Step Cut)} & \textbf{73.0 / 69.4} & 72.3 / \textbf{68.7} & \textbf{74.3 / 69.1} & \textbf{73.2 / 69.1} & \textbf{99.3} \\
    \midrule
    \rowcolor{fam}[\tabcolsep][\tabcolsep]
    \textbf{(b) Natural vs.\ Forced Cap (\%)} & \textbf{Main} & \textbf{Held-Out} & \textbf{Hard-Tail} & \textbf{\textsc{9B} Main} & \textbf{Main Finish} \\
    \midrule
    Untrained Base & 61.0 / 73.2 & 61.2 / 74.5 & 14.1 / 32.9 & 65.4 / 75.5 & 74.9\% \\
    Standard RFT (SFT) & 68.0 / 73.9 & 69.5 / \textbf{75.6} & 22.7 / 33.8 & 70.8 / \textbf{76.7} & 86.4\% \\
    On-Policy RL (\textsc{DAPO}) & 71.5 / 72.0 & 71.1 / 71.4 & 30.3 / 31.2 & -- & 98.4\% \\
    \method{} (Step Cut) & 73.7 / 73.8 & 73.2 / 73.6 & 35.3 / 35.7 & \textbf{75.5} / 75.5 & 98.8\% \\
    \rowcolor{ours}[\tabcolsep][\tabcolsep]
    \textbf{\method{} (Budget / Full$^\ast$)} & \textbf{74.7 / 74.6} & \textbf{74.3}$^\ast$ / 74.4$^\ast$ & \textbf{35.7}$^\ast$ / \textbf{35.8}$^\ast$ & -- & \textbf{100.0\%} \\
    \bottomrule
  \end{tabular}%
  }
  \vspace{-1.8mm}
  \caption{\textbf{Multi-seed reproducibility and natural vs.\ forced-cap termination across splits and scales.}
    \emph{(a)} Across $3$ independent LoRA training seeds ($29{,}000{+}$ outputs per method), \method{} (Step Cut) improves mean natural accuracy ($+5.1\%$ Main, 95\% CI $[+3.4, +6.7]$; $+3.8\%$ Reserve), tight-budget $B{=}512$ accuracy (winning on all $3$ seeds across Main and Reserve), and completion rate ($99.5\%$).
    \emph{(b)} Reporting \texttt{Natural / Forced-Cap} accuracy side by side shows that \method{} internalizes self-terminating reasoning so natural generation matches forced cap readout without external intervention ($^\ast$Full Pool on Held-Out and Hard-Tail).}
  \label{tab:multiseed-cap}
  \vspace{-2.5mm}
\end{table}

\begin{table}[t]
  \centering\footnotesize
  \setlength{\tabcolsep}{2.4pt}
  \renewcommand{\arraystretch}{0.95}
  \resizebox{\linewidth}{!}{%
  \begin{tabular}{@{}lcccccc@{}}
    \toprule
    \rowcolor{fam}[\tabcolsep][\tabcolsep]
    \textbf{Split / Method} & \textbf{$1024$} & \textbf{$1536$} & \textbf{$2048$} & \textbf{$2560$} & \textbf{$3072$} & \textbf{Cap / Tok} \\
    \midrule
    \multicolumn{7}{@{}l}{\emph{(a) Held-Out Set ($1{,}379$ Questions, Qwen3.5-4B): Accuracy (\%) / Effective Think Tok}} \\
    Untrained Base & 68.3 / 957 & 69.6 / 1357 & 70.8 / 1717 & 71.3 / 2048 & 71.8 / 2356 & 74.5 / 5820 \\
    Standard RFT (SFT) & 70.1 / 952 & 72.0 / 1331 & 73.0 / 1658 & 73.5 / 1944 & \textbf{74.1} / 2203 & \textbf{75.6} / 4523 \\
    On-Policy RL (\textsc{DAPO}) & 69.6 / \textbf{876} & 70.0 / \textbf{1137} & 70.4 / \textbf{1320} & 70.7 / \textbf{1454} & 70.7 / \textbf{1557} & 71.4 / 2023 \\
    Prefix Credit Only & 69.8 / 920 & 71.2 / 1245 & 72.0 / 1488 & 72.3 / 1655 & 72.4 / 1760 & 72.7 / 2218 \\
    \rowcolor{ours}[\tabcolsep][\tabcolsep]
    \textbf{\method{} (Full Pool)} & \textbf{71.7} / 890 & \textbf{73.3} / 1185 & \textbf{74.0} / 1402 & \textbf{73.9} / 1556 & 73.9 / 1654 & 74.4 / \textbf{1876} \\
    \midrule
    \multicolumn{7}{@{}l}{\emph{(b) Scale-Up Main Test ($1{,}200$ Questions, Qwen3.5-9B): Accuracy (\%) / Effective Think Tok}} \\
    Untrained Base (9B) & 71.6 / 939 & 72.6 / 1317 & 73.8 / 1651 & 73.8 / 1954 & 74.3 / 2231 & 75.5 / 5173 \\
    Standard RFT (9B) & \textbf{74.4} / 945 & \textbf{75.6} / 1316 & \textbf{75.7} / 1632 & \textbf{76.0} / 1909 & \textbf{76.2} / 2155 & \textbf{76.7} / 4355 \\
    \rowcolor{ours}[\tabcolsep][\tabcolsep]
    \textbf{\method{} (Matched-Row, 9B)} & 73.7 / \textbf{888} & 74.9 / \textbf{1187} & 75.0 / \textbf{1408} & 75.2 / \textbf{1568} & 75.5 / \textbf{1675} & 75.5 / \textbf{1934} \\
    \bottomrule
  \end{tabular}%
  }
  \vspace{-1.8mm}
  \caption{\textbf{Test-time thinking-budget curves on Held-Out (4B) and Main (9B).}
    \emph{(a)} On Held-Out, \method{} leads all baselines from $1{,}024$ to $2{,}560$ tokens while cutting full-cap thinking tokens by $2.41\times$ relative to Standard RFT ($4{,}523 \to 1{,}876$).
    \emph{(b)} At 9B scale, Standard RFT requires external forced readout at the cap ($76.7\%$ vs.\ $70.8\%$ natural) to recover looped rollouts, whereas \method{} achieves $75.5\%$ naturally in $1{,}934$ tokens ($99.8\%$ Finish).}
  \label{tab:frontier-heldout}
  \vspace{-2.5mm}
\end{table}

\paragraph{Reproducibility across independent training seeds.}
\Cref{tab:multiseed-cap}a and \cref{fig:supp-quant}b evaluate three independent LoRA training seeds ($17$, $29$, and $43$, with $14{,}400$ decoded outputs per method on Main and $14{,}592$ on Unseen Reserve).
Across the three seeds, \method{} (Step Cut) improves mean natural accuracy over Standard RFT by $+5.1\%$ on Main ($68.2\% \to 73.3\%$, paired 95\% CI $[+3.4, +6.7]$) and $+3.8\%$ on Reserve ($69.4\% \to 73.2\%$), while lifting completion from $86.7\%$ to $99.5\%$ ($72$ vs.\ $1{,}920$ unfinished outputs).

Under the tight $B{=}512$ budget, \method{} beats both Standard RFT and Prefix Credit Only on every seed across Main ($68.9\%$ vs.\ $65.7\%$ and $66.8\%$) and Reserve ($69.1\%$ vs.\ $67.4\%$ and $68.2\%$, paired 95\% CI $[+0.51, +1.19]$ vs.\ Prefix Credit Only across seeds; paper-cluster bootstrap $[-0.27, +1.94]$).
Moreover, $99\%$ of the Main $B{=}512$ gain over Standard RFT ($+3.25\%$ of $+3.29\%$) arises on items where both models are probed at exact token $512$.

\paragraph{Natural termination versus forced cap readout.}
\Cref{tab:multiseed-cap}b and \cref{tab:frontier-heldout} compare natural generation against forced answer readout at the $12{,}000$-token limit (\texttt{Cap}).
Because Untrained Base and Standard RFT often fail to stop ($74.9\%$ and $86.4\%$ Finish), forced cap readout raises their scores (for instance, $69.5\% \to 75.6\%$ on Held-Out and $70.8\% \to 76.7\%$ on 9B RFT), but requires $4{,}355$ to $5{,}820$ thinking tokens and external intervention.
By contrast, \method{} keeps natural and forced-cap accuracy within $0.4\%$ on every split while halving thinking tokens and leading on Hard-Tail under both natural ($+13.0\%$) and forced-cap ($+2.0\%$) evaluation.

\section{External Benchmarks and Shortcut Diagnostics}
\label{app:grounded}
\label{app:blindchannel}

\begin{figure*}[t]
  \centering
  % Supplementary Quantitative Summary (3-panel vector figure)
\resizebox{\textwidth}{!}{%
\begin{tikzpicture}[
    x=1cm, y=1cm, font=\sffamily\scriptsize,
    axisline/.style={draw=black!65, line width=0.5pt},
    gridline/.style={draw=black!14, densely dashed, line width=0.4pt},
    panelttl/.style={font=\sffamily\scriptsize\bfseries, text=hvInk}
  ]

  % ===================================================================
  % Panel (a): External Benchmark Transfer & Wall-Clock Speedup
  % ===================================================================
  \node[panelttl, anchor=west] at (0.0, 3.32) {(a) Zero-Shot External Transfer \& Speedup};
  \draw[axisline] (0.62, 0.55) rectangle (5.28, 2.82);
  \foreach \y/\ylbl in {0.55/45, 1.05/55, 1.56/65, 2.06/75, 2.57/85} {
    \draw[gridline] (0.62, \y) -- (5.28, \y);
    \node[font=\sffamily\tiny, anchor=east, text=black!80] at (0.58, \y) {\ylbl\%};
  }
  \node[font=\sffamily\tiny, rotate=90, text=black!80] at (-0.18, 1.68) {Accuracy (\%)};

  \fill[hvGrey!75, draw=hvMuted] (1.10, 0.55) rectangle (1.39, 1.86);
  \node[font=\sffamily\tiny, anchor=south, inner sep=1pt] at (1.245, 1.87) {70.9};
  \fill[qlbverm!70, draw=qlbverm!90!black] (1.45, 0.55) rectangle (1.74, 2.22);
  \node[font=\sffamily\tiny, anchor=south, inner sep=1pt] at (1.595, 2.23) {78.1};
  \fill[qlborange!80, draw=qlborange!90!black] (1.80, 0.55) rectangle (2.09, 2.06);
  \node[font=\sffamily\tiny, anchor=south, inner sep=1pt] at (1.945, 2.07) {75.0};
  \fill[qlbblue!88, draw=qlbblue!95!black] (2.15, 0.55) rectangle (2.44, 2.40);
  \node[font=\sffamily\tiny\bfseries, text=qlbblue!95!black, anchor=south, inner sep=1pt] at (2.295, 2.41) {81.8};
  \node[font=\sffamily\tiny\bfseries, fill=hvMint, draw=hvTeal!55, rounded corners=1.5pt, inner xsep=2pt, inner ysep=0.8pt] at (1.36, 2.66) {3.28$\times$ faster};
  \node[font=\sffamily\tiny\bfseries, anchor=north, text=black!85] at (1.77, 0.48) {MathVista (1,000)};

  \fill[hvGrey!75, draw=hvMuted] (3.34, 0.55) rectangle (3.63, 1.17);
  \node[font=\sffamily\tiny, anchor=south, inner sep=1pt] at (3.485, 1.18) {57.3};
  \fill[qlbverm!70, draw=qlbverm!90!black] (3.69, 0.55) rectangle (3.98, 1.29);
  \node[font=\sffamily\tiny, anchor=south, inner sep=1pt] at (3.835, 1.30) {59.7};
  \fill[qlborange!80, draw=qlborange!90!black] (4.04, 0.55) rectangle (4.33, 1.29);
  \node[font=\sffamily\tiny, anchor=south, inner sep=1pt] at (4.185, 1.30) {59.6};
  \fill[qlbblue!88, draw=qlbblue!95!black] (4.39, 0.55) rectangle (4.68, 1.74);
  \node[font=\sffamily\tiny\bfseries, text=qlbblue!95!black, anchor=south, inner sep=1pt] at (4.535, 1.75) {68.6};
  \node[font=\sffamily\tiny\bfseries, fill=hvMint, draw=hvTeal!55, rounded corners=1.5pt, inner xsep=2pt, inner ysep=0.8pt] at (4.01, 2.15) {2.47$\times$ faster};
  \node[font=\sffamily\tiny\bfseries, anchor=north, text=black!85] at (4.01, 0.48) {ChartQA (2,500)};

  \fill[hvGrey!75, draw=hvMuted] (0.64, 2.94) rectangle (0.78, 3.06);
  \node[font=\sffamily\tiny, anchor=west] at (0.81, 3.00) {Base};
  \fill[qlbverm!70, draw=qlbverm!90!black] (1.52, 2.94) rectangle (1.66, 3.06);
  \node[font=\sffamily\tiny, anchor=west] at (1.69, 3.00) {Std.\ RFT};
  \fill[qlborange!80, draw=qlborange!90!black] (2.70, 2.94) rectangle (2.84, 3.06);
  \node[font=\sffamily\tiny, anchor=west] at (2.87, 3.00) {DAPO};
  \fill[qlbblue!88, draw=qlbblue!95!black] (3.66, 2.94) rectangle (3.80, 3.06);
  \node[font=\sffamily\tiny\bfseries, text=qlbblue!95!black, anchor=west] at (3.83, 3.00) {\method{} (Budget)};

  % ===================================================================
  % Panel (b): Three-Seed Reproducibility (Main Natural & B=512)
  % ===================================================================
  \node[panelttl, anchor=west] at (5.75, 3.32) {(b) Stability Across 3 Training Seeds};
  \draw[axisline] (6.32, 0.55) rectangle (10.85, 2.82);
  \foreach \y/\ylbl in {0.55/63, 1.09/66, 1.64/69, 2.18/72, 2.73/75} {
    \draw[gridline] (6.32, \y) -- (10.85, \y);
    \node[font=\sffamily\tiny, anchor=east, text=black!80] at (6.28, \y) {\ylbl\%};
  }
  \foreach \x/\slbl in {7.05/Seed 17, 8.58/Seed 29, 10.10/Seed 43} {
    \draw[axisline] (\x, 0.55) -- (\x, 0.49);
    \node[font=\sffamily\tiny\bfseries, anchor=north, text=black!85] at (\x, 0.48) {\slbl};
  }
  \node[font=\sffamily\tiny, text=black!80] at (8.58, 0.05) {Independent LoRA runs (4,800 outputs / seed)};

  \draw[qlbblue, line width=1.25pt] (7.05, 2.49) -- (8.58, 2.29) -- (10.10, 2.47);
  \foreach \x/\y/\v in {7.05/2.49/73.7, 8.58/2.29/72.6, 10.10/2.47/73.6} {
    \fill[qlbblue] (\x, \y) circle (1.9pt);
    \node[font=\sffamily\tiny\bfseries, text=qlbblue!95!black, anchor=south] at (\x, {\y+0.03}) {\v};
  }
  \draw[qlbgreen!85!black, densely dashed, line width=1.05pt] (7.05, 1.60) -- (8.58, 1.62) -- (10.10, 1.67);
  \foreach \x/\y/\v in {7.05/1.60/68.8, 8.58/1.62/68.9, 10.10/1.67/69.2} {
    \fill[qlbgreen!85!black] (\x, \y) circle (1.6pt);
    \node[font=\sffamily\tiny\bfseries, text=qlbgreen!80!black, anchor=south] at (\x, {\y+0.03}) {\v};
  }
  \draw[qlbverm, line width=1.1pt] (7.05, 1.46) -- (8.58, 1.50) -- (10.10, 1.54);
  \foreach \x/\y/\v in {7.05/1.46/68.0, 8.58/1.50/68.2, 10.10/1.54/68.5} {
    \fill[qlbverm] (\x, \y) circle (1.7pt);
  }
  \node[font=\sffamily\tiny, text=qlbverm!90!black, anchor=north] at (7.05, 1.42) {68.0};
  \node[font=\sffamily\tiny, text=qlbverm!90!black] at (8.58, 1.35) {68.2};
  \node[font=\sffamily\tiny, text=qlbverm!90!black, anchor=north] at (10.10, 1.50) {68.5};

  \draw[qlbverm!80, densely dashed, line width=0.95pt] (7.05, 0.87) -- (8.58, 1.19) -- (10.10, 1.04);
  \foreach \x/\y/\v in {7.05/0.87/64.8, 8.58/1.19/66.5, 10.10/1.04/65.7} {
    \fill[white, draw=qlbverm, line width=0.8pt] (\x, \y) circle (1.5pt);
    \node[font=\sffamily\tiny, text=qlbverm!85!black, anchor=north] at (\x, {\y-0.03}) {\v};
  }

  \draw[qlbblue, line width=1.2pt] (6.36, 3.00) -- (6.62, 3.00);
  \fill[qlbblue] (6.49, 3.00) circle (1.6pt);
  \node[font=\sffamily\tiny\bfseries, text=qlbblue!95!black, anchor=west] at (6.65, 3.00) {\method{} (Nat.)};
  \draw[qlbverm, line width=1.1pt] (8.05, 3.00) -- (8.31, 3.00);
  \fill[qlbverm] (8.18, 3.00) circle (1.6pt);
  \node[font=\sffamily\tiny, text=qlbverm!90!black, anchor=west] at (8.34, 3.00) {RFT (Nat.)};
  \draw[qlbgreen!85!black, densely dashed, line width=1.0pt] (9.55, 3.00) -- (9.81, 3.00);
  \fill[qlbgreen!85!black] (9.68, 3.00) circle (1.5pt);
  \node[font=\sffamily\tiny\bfseries, text=qlbgreen!80!black, anchor=west] at (9.84, 3.00) {$B{=}512$};

  % ===================================================================
  % Panel (c): Hard-Tail Modality Diagnostic (Answer-Only vs. PRFT)
  % ===================================================================
  \node[panelttl, anchor=west] at (11.35, 3.32) {(c) Input Modality on Hard-Tail \& Main};
  \draw[axisline] (11.92, 0.55) rectangle (16.45, 2.82);
  \foreach \y/\ylbl in {0.55/20, 1.31/40, 2.06/60, 2.82/80} {
    \draw[gridline] (11.92, \y) -- (16.45, \y);
    \node[font=\sffamily\tiny, anchor=east, text=black!80] at (11.88, \y) {\ylbl\%};
  }

  \fill[qlbblue!88, draw=qlbblue!95!black] (12.24, 0.55) rectangle (12.58, 2.58);
  \node[font=\sffamily\tiny\bfseries, text=qlbblue!95!black, anchor=south, inner sep=1pt] at (12.41, 2.59) {73.7};
  \fill[qlbblue!28, draw=qlbblue!85!black] (12.62, 0.55) rectangle (12.96, 1.20);
  \node[font=\sffamily\tiny, text=qlbblue!95!black, anchor=south, inner sep=1pt] at (12.79, 1.21) {37.1};
  \node[font=\sffamily\tiny\bfseries, anchor=north, align=center] at (12.60, 0.48) {\method{}\\{\tiny(Main)}};

  \fill[qlbverm!78, draw=qlbverm!90!black] (13.52, 0.55) rectangle (13.86, 2.62);
  \node[font=\sffamily\tiny, text=qlbverm!90!black, anchor=south, inner sep=1pt] at (13.69, 2.63) {74.8};
  \fill[qlbverm!28, draw=qlbverm!85!black] (13.90, 0.55) rectangle (14.24, 1.70);
  \node[font=\sffamily\tiny\bfseries, text=qlbverm!90!black, anchor=south, inner sep=1pt] at (14.07, 1.71) {50.4};
  \node[font=\sffamily\tiny\bfseries, anchor=north, align=center] at (13.88, 0.48) {Ans.-Only\\{\tiny(Main)}};

  \fill[qlbverm!78, draw=qlbverm!90!black] (14.84, 0.55) rectangle (15.16, 1.70);
  \node[font=\sffamily\tiny, text=qlbverm!90!black, anchor=south, inner sep=1pt] at (15.00, 1.71) {50.4};
  \fill[qlbverm!50, draw=qlbverm!85!black] (15.20, 0.55) rectangle (15.52, 1.65);
  \node[font=\sffamily\tiny, text=qlbverm!90!black, anchor=south, inner sep=1pt] at (15.36, 1.66) {49.1};
  \fill[qlbverm!25, draw=qlbverm!85!black] (15.56, 0.55) rectangle (15.88, 1.65);
  \node[font=\sffamily\tiny\bfseries, text=qlbverm!90!black, anchor=south, inner sep=1pt] at (15.72, 1.66) {49.0};
  \node[font=\sffamily\tiny\bfseries, fill=hvSand, draw=hvAmber!55, rounded corners=1.5pt, inner xsep=2pt, inner ysep=0.8pt] at (15.36, 2.18) {$-1.4\%$ w/o img};
  \node[font=\sffamily\tiny\bfseries, anchor=north, align=center] at (15.36, 0.48) {Ans.-Only\\{\tiny(Hard-Tail)}};

  \fill[black!72] (11.96, 2.94) rectangle (12.10, 3.06);
  \node[font=\sffamily\tiny, anchor=west] at (12.13, 3.00) {Img.+Cap.};
  \fill[black!42] (13.40, 2.94) rectangle (13.54, 3.06);
  \node[font=\sffamily\tiny, anchor=west] at (13.57, 3.00) {Cap.\ Only};
  \fill[black!18, draw=black!65] (14.80, 2.94) rectangle (14.94, 3.06);
  \node[font=\sffamily\tiny, anchor=west] at (14.97, 3.00) {Neither};

\end{tikzpicture}%
}
  \vspace{-2.5mm}
  \caption{\textbf{Supplementary quantitative diagnostics across external benchmarks, training seeds, and input modalities.}
    \emph{(a)} In zero-shot transfer to \textsc{MathVista} and \textsc{ChartQA}, \method{} (Budget Cut) improves accuracy over Base, Standard RFT, and \textsc{DAPO} while accelerating batch evaluation by $3.28\times$ and $2.47\times$ over Standard RFT.
    \emph{(b)} Across three independent LoRA training seeds, \method{} (Step Cut) consistently outperforms Standard RFT under both natural termination and a tight $512$-token budget.
    \emph{(c)} Removing both the figure and caption (\texttt{Neither}) causes a $36.6\%$ drop for \method{} on Main, whereas \texttt{Answer-Only SFT} retains $\sim 49\%$ to $50\%$ accuracy on both Main and Hard-Tail without seeing the figure.}
  \label{fig:supp-quant}
  \vspace{-3.0mm}
\end{figure*}
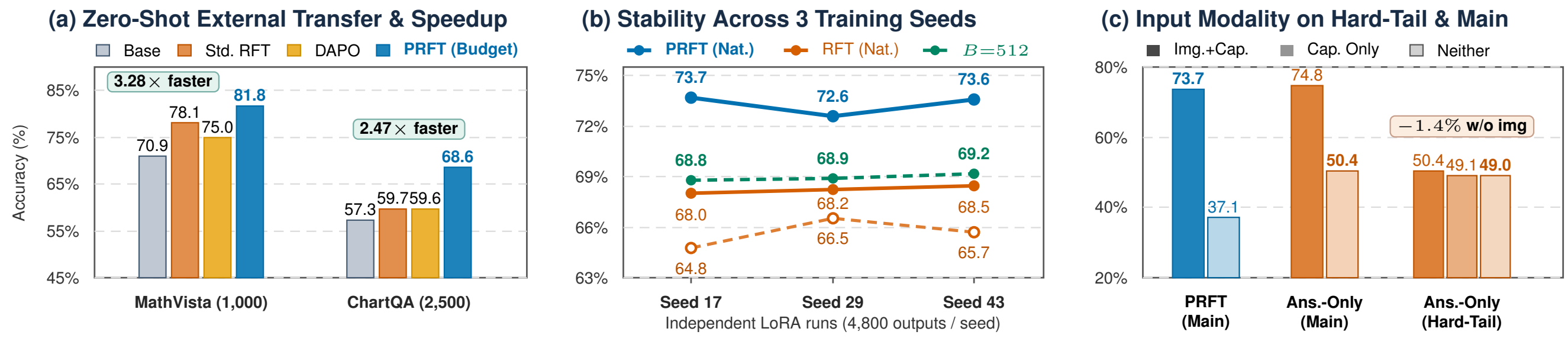

\begin{table*}[t]
  \centering\footnotesize
  \setlength{\tabcolsep}{3.4pt}
  \renewcommand{\arraystretch}{0.96}
  \resizebox{\textwidth}{!}{%
  \begin{tabular}{@{}lcccccccccccc@{}}
    \toprule
    & \multicolumn{5}{c}{\textbf{\textsc{MathVista} (\texttt{testmini}, $1{,}000$)}} & \multicolumn{4}{c}{\textbf{\textsc{ChartQA} (\texttt{test}, $2{,}500$)}} & \multicolumn{3}{c}{\textbf{\textsc{MathVision} (\texttt{testmini}, $304$)}} \\
    \cmidrule(lr){2-6}\cmidrule(lr){7-10}\cmidrule(lr){11-13}
    \textbf{Method (\textsc{Qwen3.5-4B})} & \textbf{All} & \textbf{MC} & \textbf{Free} & \textbf{Finish} & \textbf{Tok / Time} & \textbf{All} & \textbf{Hum. / Aug.} & \textbf{Finish} & \textbf{Tok / Time} & \textbf{All} & \textbf{MC / Free} & \textbf{Finish / Tok} \\
    \midrule
    Untrained Base & 70.9 & 70.7 & 71.1 & 88.9\% & 3,543 / 42.4m & 57.3 & 65.5 / 49.1 & 97.3\% & 1,272 / 73.8m & 37.5 & 36.8 / 38.6 & 41.8\% / 9,028 \\
    Standard RFT (SFT) & 78.1 & 82.2 & 73.3 & 91.2\% & 2,909 / 55.0m & 59.7 & 65.9 / 53.5 & 98.7\% & 1,015 / 87.2m & 46.7 & 45.8 / 48.3 & 56.6\% / 8,065 \\
    On-Policy RL (\textsc{DAPO}) & 75.0 & 74.6 & \textbf{75.4} & 97.3\% & 1,801 / 48.7m & 59.6 & 67.2 / 51.9 & \textbf{100.0\%} & \textbf{541} / 38.1m & \textbf{54.0} & \textbf{54.7 / 52.6} & 73.7\% / 6,467 \\
    \rowcolor{ours}[\tabcolsep][\tabcolsep]
    \method{} (Step Cut) & 79.2 & 84.3 & 73.3 & 99.3\% & 1,758 / 37.6m & 60.4 & 64.9 / 55.9 & 99.9\% & 790 / 53.6m & 47.7 & 49.5 / 44.7 & 96.4\% / 2,941 \\
    \rowcolor{ours}[\tabcolsep][\tabcolsep]
    \textbf{\method{} (Budget Cut)} & \textbf{81.8} & \textbf{87.8} & 74.8 & \textbf{100.0\%} & \textbf{1,710 / 16.7m} & \textbf{68.6} & \textbf{69.8 / 67.5} & \textbf{100.0\%} & 784 / \textbf{35.3m} & 48.4 & 53.2 / 40.4 & \textbf{100.0\% / 2,842} \\
    \bottomrule
  \end{tabular}%
  }
  \vspace{-1.8mm}
  \caption{\textbf{Zero-shot transfer to external visual reasoning benchmarks (single decoding seed $2001$).}
    Evaluated on \textsc{MathVista} ($540$ multiple-choice \textbf{MC} and $460$ open-ended \textbf{Free}), \textsc{ChartQA} ($1{,}250$ \textbf{Hum.} and $1{,}250$ \textbf{Aug.} open-ended chart questions), and \textsc{MathVision} ($190$ \textbf{MC} and $114$ \textbf{Free} competition math items on B200).
    \method{} (Budget Cut) achieves $100\%$ completion across all three benchmarks, outperforms SFT and RL on \textsc{MathVista} ($81.8\%$, $3.28\times$ faster than RFT) and \textsc{ChartQA} ($68.6\%$, $2.47\times$ faster than RFT), and beats Standard RFT overall on \textsc{MathVision} ($48.4\%$ vs.\ $46.7\%$) at $2.84\times$ fewer thinking tokens, while \textsc{DAPO} leads on long open-ended \textsc{MathVision} derivations.}
  \label{tab:external-benchmarks}
  \vspace{-2.5mm}
\end{table*}

\begin{table}[t]
  \centering\footnotesize
  \setlength{\tabcolsep}{2.8pt}
  \renewcommand{\arraystretch}{0.95}
  \resizebox{\linewidth}{!}{%
  \begin{tabular}{@{}lccccc@{}}
    \toprule
    \rowcolor{fam}[\tabcolsep][\tabcolsep]
    \textbf{Split / Method} & \textbf{Gold A} & \textbf{Gold B} & \textbf{Gold C} & \textbf{Gold D} & \textbf{Option Bias} \\
    \midrule
    \multicolumn{6}{@{}l}{\emph{(a) Main Test ($1{,}200$ Questions; Gold Counts: $283$ / $312$ / $309$ / $296$)}} \\
    Prefix Credit Only & 70.1\% & 71.1\% & 71.4\% & 72.7\% & \textbf{2.7\%} \\
    \rowcolor{ours}[\tabcolsep][\tabcolsep]
    \textbf{\method{} (Step Cut)} & \textbf{72.4\%} & \textbf{73.8\%} & \textbf{73.1\%} & \textbf{75.6\%} & 3.2\% \\
    Answer-Only SFT (Zero-Think Pool) & 56.5\% & 81.8\% & 68.4\% & 71.1\% & 25.3\% \\
    Answer-Only SFT (w/o Prefix) & 59.3\% & 84.7\% & 73.2\% & 80.7\% & \textcolor{qlbverm}{\textbf{25.4\%}} \\
    \midrule
    \multicolumn{6}{@{}l}{\emph{(b) Hard-Tail Test ($2{,}306$ Questions; Gold Counts: $630$ / $603$ / $533$ / $540$)}} \\
    On-Policy RL (\textsc{DAPO}) & 27.5\% & 33.4\% & 30.1\% & 30.4\% & \textbf{5.9\%} \\
    Prefix Credit Only & 28.4\% & 36.3\% & 31.8\% & 32.2\% & 7.9\% \\
    \rowcolor{ours}[\tabcolsep][\tabcolsep]
    \textbf{\method{} (Step Cut)} & \textbf{32.1\%} & \textbf{38.6\%} & \textbf{35.0\%} & \textbf{35.6\%} & 6.5\% \\
    Answer-Only SFT (w/o Prefix) & 25.0\% & 64.0\% & 50.1\% & 63.9\% & \textcolor{qlbverm}{\textbf{39.0\%}} \\
    \bottomrule
  \end{tabular}%
  }
  \vspace{-1.8mm}
  \caption{\textbf{Per-option accuracy across gold answer letters A to D.}
    \textbf{Option Bias} measures the max-to-min accuracy gap across gold options A--D.
    Supervising the reasoning prefix in \method{} keeps accuracy balanced across options ($3.2\%$ and $6.5\%$ bias), whereas stripping the prefix (\texttt{Answer-Only SFT}) induces severe option-letter bias ($25.4\%$ and $39.0\%$).}
  \label{tab:letter-spread}
  \vspace{-2.5mm}
\end{table}

\paragraph{Zero-shot protocol on external benchmarks.}
\Cref{tab:external-benchmarks} and \cref{fig:supp-quant}a evaluate five fixed \textsc{Qwen3.5-4B} checkpoints zero-shot on \textsc{MathVista}~\cite{mathvista2024}, \textsc{ChartQA}~\cite{chartqa2022}, and \textsc{MathVision}~\cite{wang2024} under identical thinking-mode decoding ($T=1.0$, $\text{top-}p=0.95$, $\text{top-}k=20$, presence penalty $1.5$, $12{,}000$-token cap, single seed $2001$; note this differs from greedy leaderboard decoding).
Predictions are scored with deterministic JSON/boxed/final-answer extraction and author normalization on \textsc{MathVista} and \textsc{MathVision} (without an LLM extractor) and standard $5\%$ relaxed numerical matching on \textsc{ChartQA}.

On \textsc{MathVista} and \textsc{ChartQA}, \method{} (Budget Cut) reaches \textbf{81.8\%} and \textbf{68.6\%} with $100\%$ completion and $<0.7\%$ parse failures (vs.\ $11.0\%$ on \textsc{MathVista} for RFT), running \textbf{$3.28\times$} and \textbf{$2.47\times$ faster} than RFT.
Post-hoc lexical inspection shows that part of the \textsc{ChartQA} margin reflects cleaner output formatting (among $340$ items turning from RFT wrong to \method{} right, $94$ RFT predictions appended a trailing \texttt{\%} to the gold integer and $70$ differed only by spaces or commas), alongside reduced truncation and parse errors on \textsc{MathVista}.
On competition-level \textsc{MathVision}, \method{} beats Base ($+10.9\%$) and Standard RFT ($+1.7\%$ overall, $+7.4\%$ on MC) at $2.84\times$ fewer tokens, while \textsc{DAPO} leads on open-ended symbolic derivations ($52.6\%$ vs.\ $40.4\%$) that exceed our $3{,}072$-token prefix budget (\cref{fig:supp-failure}c).

\paragraph{Shortcut diagnostics for Answer-Only SFT.}
\Cref{tab:letter-spread} and \cref{fig:supp-quant}c examine why \texttt{Answer-Only SFT} (trained on the \method{}-selected pool with answer labels only) scores $74.8\%$ on Main and $50.1\%$ on Hard-Tail with zero thinking tokens.
First, stripping the prefix induces severe imbalance across options (\cref{tab:letter-spread}), scoring only $25.0\%$ on Gold A versus $64.0\%$ on Gold B on Hard-Tail ($39.0\%$ Option Bias, despite A being the most frequent label at $630/2{,}306$ items).
Second, in a paired $27{,}672$-sample modality test on Hard-Tail ($2{,}306$ items $\times$ $4$ seeds $\times$ $3$ input modes), \texttt{Answer-Only SFT} reaches $50.4\%$ with image and caption, $49.1\%$ with caption only, and \textbf{49.0\% with question and options alone} ($-1.4\%$ without the figure, though $29.4\%$ of individual predictions still flip).
Non-visual question-option cues thus account for nearly its entire Hard-Tail average score, whereas prefix supervision in \method{} keeps Option Bias low ($3.2\%$ Main, $6.5\%$ Hard-Tail) and preserves strong image dependence (\cref{tab:modality-grid}b).

\section{Teacher Probing, Latency, and Gradients}
\label{app:capsweep}
\label{app:latency}
\label{app:theta}

\begin{figure}[t]
  \centering
  \includegraphics[width=0.81\linewidth]{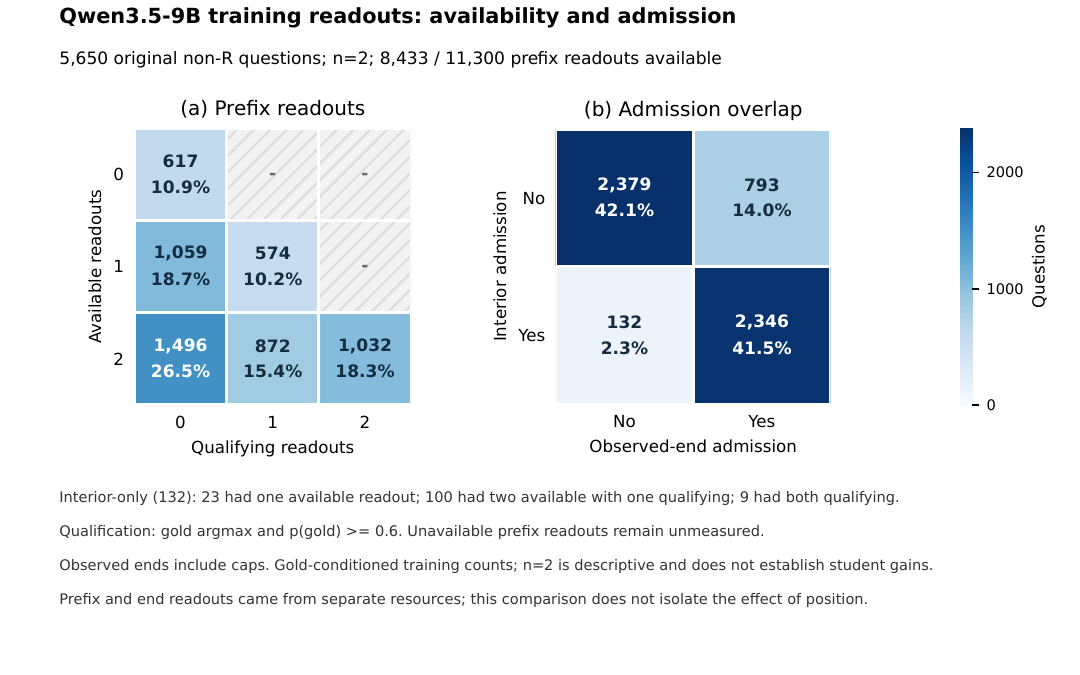}
  \vspace{-3.5mm}
  \caption{\textbf{Readout census on $5{,}650$ unfinished \textsc{Qwen3.5-9B} questions ($11{,}300$ rollouts).}
    \emph{(a)} Step-boundary prefix readout availability across $n{=}2$ rollouts ($8{,}433 / 11{,}300$ contain a break in $[256, 3{,}072]$).
    \emph{(b)} Overlap with endpoint forced readout: $2{,}346$ pass at both, $132$ only at the prefix, and fixed-$3{,}072$ probing rescues $701$ additional questions without early breaks.}
  \label{fig:teacher9b-census}
  \vspace{-3.5mm}
\end{figure}

\begin{table}[t]
  \centering\footnotesize
  \setlength{\tabcolsep}{3.0pt}
  \renewcommand{\arraystretch}{0.92}
  \resizebox{\linewidth}{!}{%
  \begin{tabular}{@{}lcccccccc@{}}
    \toprule
    \rowcolor{fam}[\tabcolsep][\tabcolsep]
    \textbf{Readout Threshold} & $\mathbf{0.0}$ & $\mathbf{0.3}$ & $\mathbf{0.5}$ & $\mathbf{0.6}$ & $\mathbf{0.7}$ & $\mathbf{0.8}$ & $\mathbf{0.9}$ & $\mathbf{0.99}$ \\
    \midrule
    Budget Cut Pass Rate (\%) & \textbf{85.9} & \textbf{85.9} & 85.2 & 84.2 & 83.1 & 81.8 & 79.8 & 75.4 \\
    Step Cut Pass Rate (\%) & 81.5 & 81.5 & 80.5 & 79.3 & 78.0 & 76.5 & 74.7 & 71.5 \\
    RFT-Admitted Missed & 0 & 0 & 0 & 0 & 0 & 0 & 0 & 0 \\
    \bottomrule
  \end{tabular}%
  }
  \vspace{-2.2mm}
  \caption{\textbf{Teacher pass-rate sensitivity to readout confidence threshold} on the $9{,}702$-question \textsc{Qwen3.5-27B} single-cap pool ($n{=}2$).
    Because correct prefix readouts are near-deterministic (median confidence $1.000$), pass rate varies by only $4.1\%$ across $[0.0, 0.8]$.}
  \label{tab:theta-sweep}
  \vspace{-3.2mm}
\end{table}

\begin{table}[t]
  \centering\footnotesize
  \setlength{\tabcolsep}{2.2pt}
  \renewcommand{\arraystretch}{0.92}
  \resizebox{\linewidth}{!}{%
  \begin{tabular}{@{}lccccc@{}}
    \toprule
    \rowcolor{fam}[\tabcolsep][\tabcolsep]
    \multicolumn{6}{@{}l}{\textbf{(a) Live 4$\times$H100 Student Inference (64-Query Panel) \& Online Teacher Probing}} \\
    \textbf{Pipeline / Model Arm} & \textbf{Tok / q} & \textbf{Wall (s)} & \textbf{Latency / q} & \textbf{Speedup / Cost} & \textbf{Acc.\ / Finish} \\
    \midrule
    Untrained Base (4B) & 5,622 & 367.6s & 5,743 ms & $1.12\times$ & 59.0\% / 77.3\% \\
    Standard RFT (4B) & 4,614 & 412.9s & 6,451 ms & $1.00\times$ (Base) & 67.6\% / 89.8\% \\
    Prefix Credit Only (4B) & \textbf{2,099} & \textbf{329.3s} & \textbf{5,145 ms} & \textbf{1.25$\times$ faster} & 70.7\% / \textbf{99.2\%} \\
    \rowcolor{ours}[\tabcolsep][\tabcolsep]
    \textbf{\method{} (Step Cut, 4B)} & 2,423 & 361.8s & 5,653 ms & \textbf{1.14$\times$ faster} & \textbf{73.8\%} / 97.3\% \\
    \cmidrule(lr){1-6}
    $8\times$ Endpoint Resampling (27B) & -- & 827.7s & -- & $3.75\times$ time & 60.9\% Pass \\
    \rowcolor{ours}[\tabcolsep][\tabcolsep]
    \textbf{\method{} ($1\times$ + Warm KV Cache)} & -- & \textbf{218.9 + 1.7s} & -- & \textbf{+0.78\% time} & \textbf{64.1\% Pass} \\
    \midrule
    \rowcolor{fam}[\tabcolsep][\tabcolsep]
    \multicolumn{6}{@{}l}{\textbf{(b) Final-Checkpoint Local LoRA Gradients ($144$ Records = $48$ Targets $\times$ $3$ Seeds)}} \\
    \textbf{Supervision Target Group} & \textbf{Med.\ Tok} & \textbf{Prefix CE} & \textbf{Stop CE} & \textbf{$\lVert g_{\mathrm{pre}}\rVert_2$} & \textbf{Med.\ $\cos(g_{\mathrm{tot}}, g_{\mathrm{stop}})$} \\
    \midrule
    Natural Finished ($n{=}48$) & 687.0 & 0.161 & $9.91\times 10^{-2}$ & 1.351 & \textbf{+0.332} \\
    Shared Prefix-Cut ($n{=}48$) & 2,534.5 & 0.213 & \textbf{5.89$\times$10$^{-5}$} & 0.996 & +0.011 \\
    Rescued Prefix-Cut ($n{=}48$) & 2,450.5 & 0.223 & \textbf{5.79$\times$10$^{-5}$} & 0.932 & +0.014 \\
    \bottomrule
  \end{tabular}%
  }
  \vspace{-2.2mm}
  \caption{\textbf{H100 inference latency, online probing cost, and local LoRA gradient geometry.}
    \emph{(a)} On $4\times\text{H100}$ GPUs (4-round rotation, $64$ queries), \method{} speeds up generation by $1.14\times$--$1.25\times$, while warm-cache probing ($16$-query panel) adds $0.78\%$ wall time ($3.75\times$ faster than $8\times$ resampling).
    \emph{(b)} Pooled 3-seed medians at final checkpoints show near-zero stop-suffix CE ($\sim 6\times 10^{-5}$) on cut targets, leaving $\approx 99\%$ of local gradient norm on the prefix.}
  \label{tab:h100-latency}
  \label{tab:audit-grad}
  \vspace{-3.2mm}
\end{table}

\paragraph{Unfinished teacher census and threshold stability.}
On the $11{,}702$-question \textsc{Qwen3.5-27B} pool, endpoint RFT admits $6{,}656$ items ($56.9\%$ Pass Rate, with $>97\%$ of rejected rollouts hitting the token limit), whereas \method{} admits $9{,}366$ items ($80.0\%$, rescuing $2{,}740$ unfinished items; $30$ endpoint-only items arise solely from the $2{,}000$-item $8{,}192$-cap slice).
On the $1{,}627$-question Hard-Tail Pool ($n{=}8$), \method{} rescues $671$ items ($41.2\%$) vs.\ $93$ ($5.7\%$) for $8\times$ resampling~\cite{tong2024dart} ($7.22\times$ gain, \cref{fig:overview}c).

\Cref{fig:teacher9b-census} inspects all $5{,}650$ endpoint-rejected items ($11{,}300$ rollouts) of \textsc{Qwen3.5-9B}: among $2{,}478$ prefix-rescued items, continuing to token $8{,}192$ causes $211$ selected rollouts to fail endpoint readout ($120$ shift to a non-gold argmax and $91$ drop below $0.60$ confidence), while fixed-$3{,}072$ probing rescues $701$ extra items without early paragraph breaks.
\Cref{tab:theta-sweep} confirms that pass rates are insensitive to confidence thresholds, while prefix truncation reduces SFT response tokens by $1.48\times$ and quadratic padded-length cost by $1.73\times$ over full-length endpoint targets.

\paragraph{Hardware latency, training compute, and local gradients.}
On $4\times\text{NVIDIA H100 80GB}$ GPUs (\texttt{vLLM 0.19.1}, 4-round Latin-square rotation; \cref{tab:h100-latency}a), halving student output tokens yields a $1.14\times$ (\method{} Step Cut) to $1.25\times$ (Prefix Credit Only) batch generation speedup.
During online teacher collection ($16$-question warm-cache panel), probing inside the active vLLM engine hits the prefix cache $82\%$ to $88\%$ of the time and adds only $0.78\%$ median wall-clock time ($3.75\times$ faster than $8\times$ resampling).

In offline compute accounting, generating $23{,}404$ \textsc{Qwen3.5-27B} rollouts takes $4.13$ GPU-hours, standalone offline prefix probing across $9{,}578$ unfinished rollouts takes $0.52$ GPU-hours ($12.7\%$ of rollout time without warm-engine reuse), and 3-epoch 4B LoRA SFT takes $7.94$ to $10.07$ GPU-hours (vs.\ $10.54$ GPU-hours plus $1.47\text{ GiB}$ of cached top-$16$ logits for \textsc{OPD}).
Evaluating local LoRA gradients on $48$ fixed targets across three final \method{} checkpoints ($144$ records, \cref{tab:audit-grad}b) shows that the stop suffix attains near-zero loss ($\sim 6\times 10^{-5}$) on cut traces, leaving roughly $99\%$ of the local gradient norm on the prefix (note this is a final-checkpoint diagnostic rather than an optimization trajectory).

\section{Additional Success and Failure Case Studies}
\label{app:cases}
\label{app:limitations}

\begin{figure*}[p]
  \centering
  % Supplementary Qualitative Success Cases (3 verified visual cases)
\providecommand{\suptileok}[1]{\colorbox{qlbgreen!85!black}{\textcolor{white}{\texttt{\textbf{#1}}}}}
\providecommand{\suptileno}[1]{\colorbox{qlbverm!18}{\textcolor{qlbverm!90!black}{\texttt{#1}}}}
\resizebox{\textwidth}{!}{%
\setlength{\fboxsep}{1.1pt}%
\setlength{\tabcolsep}{4.5pt}%
\renewcommand{\arraystretch}{1.08}%
\begin{tabular}{@{}p{0.322\textwidth}p{0.322\textwidth}p{0.322\textwidth}@{}}
  \toprule
  \textbf{(a) Zero-Shot External Chart Transfer} &
  \textbf{(b) Curve-Offset Deduction ($12$ Runs)} &
  \textbf{(c) Unseen Spectrum Plot ($12$ Runs)} \\
  \scriptsize \textsc{ChartQA} Test (Open-Ended Numerical) &
  \scriptsize \textsc{VeriSciQA} Main Test (Nuclear Physics) &
  \scriptsize \textsc{VeriSciQA} Unseen Reserve (Quantum Physics) \\
  \midrule
  \multicolumn{1}{c}{\includegraphics[height=2.55cm,keepaspectratio]{figures/cases/case_chartqa_87.png}} &
  \multicolumn{1}{c}{\includegraphics[height=2.55cm,keepaspectratio]{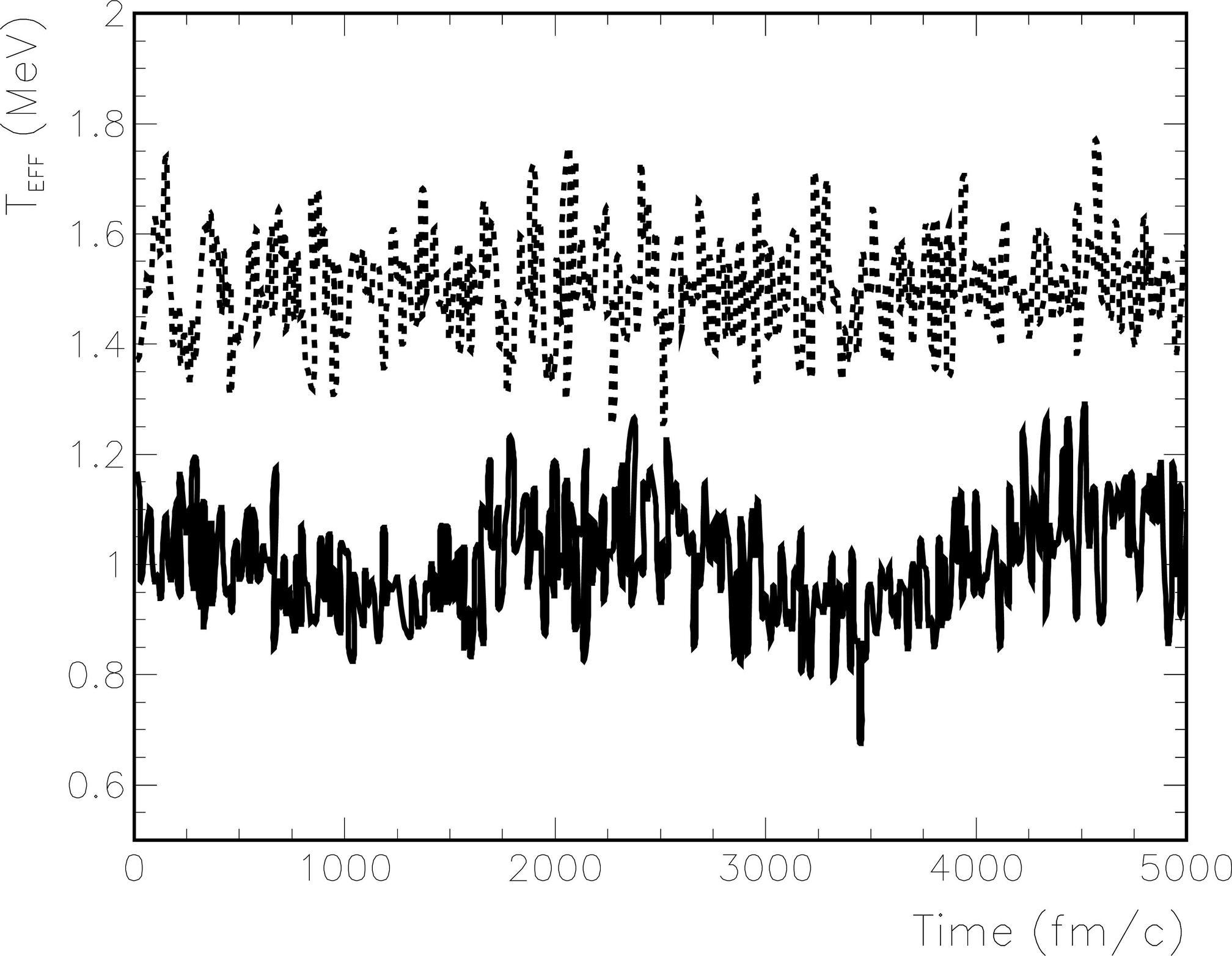}} &
  \multicolumn{1}{c}{\includegraphics[height=2.55cm,keepaspectratio]{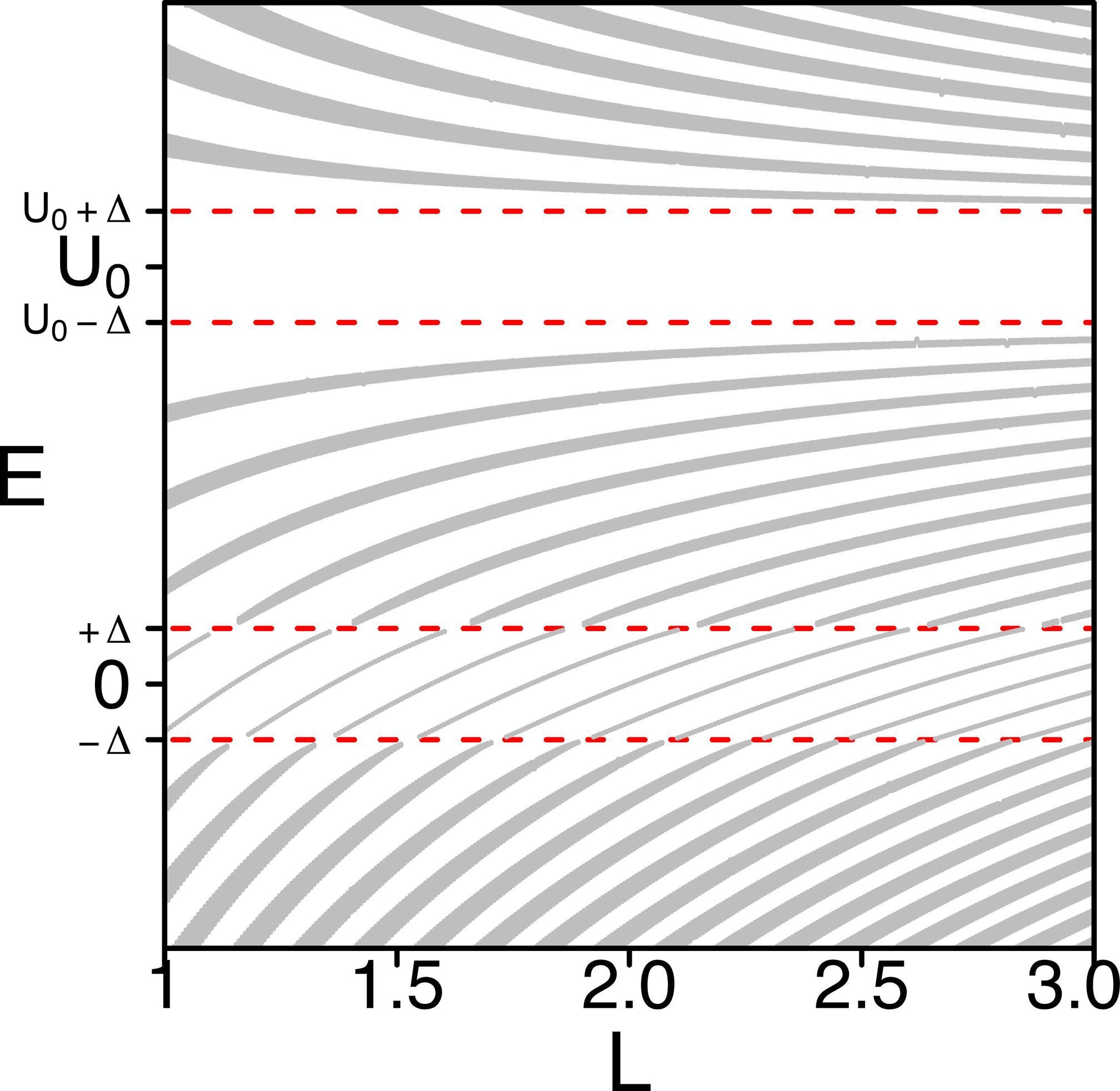}} \\
  \midrule
  \scriptsize \textbf{Q:} What is the sum of the three same values in the lower graph? (\textbf{Gold: 141}, $47{+}47{+}47$) &
  \scriptsize \textbf{Q:} What is the magnitude of the offset added to the upper curve in the figure? (\textbf{Gold: D}, $0.5\text{ MeV}$) &
  \scriptsize \textbf{Q:} Which energy region contains neither bound states nor QBSs for $L \in [1, 3]$? (\textbf{Gold: A}, $U_0{-}\Delta \le E \le U_0{+}\Delta$) \\
  \midrule
  \scriptsize \textcolor{qlbverm}{\textbf{Standard RFT (Wrong, 10,848 tok):}} Identifies the lower line early, then second-guesses repeated labels for over $10\text{k}$ tokens: \emph{``the 36s form a clearer set of three values\dots\ $36{+}36{+}36 = 108$''} (\texttt{"108"}). &
  \scriptsize \textcolor{qlbverm}{\textbf{Standard RFT ($8/12$ Nat., $4{,}505$ tok):}} Drifts between peak-to-peak and mean band offsets.\par\vspace{1.8pt}
  \scriptsize \textbf{$B{=}512$ ($8/12$):} \suptileno{A}\suptileno{A}\suptileok{D}\suptileno{A}\,\suptileok{D}\suptileok{D}\suptileok{D}\suptileno{A}\,\suptileok{D}\suptileok{D}\suptileok{D}\suptileok{D}\par\vspace{1.8pt}
  \scriptsize \textbf{Prefix Credit Only ($5/12$):} \suptileok{D}\suptileok{D}\suptileno{A}\suptileno{B}\,\suptileok{D}\suptileok{D}\suptileno{A}\suptileno{C}\,\suptileno{C}\suptileno{C}\suptileno{C}\suptileok{D} &
  \scriptsize \textcolor{qlbverm}{\textbf{Standard RFT ($8/12$ Nat., $7{,}216$ tok):}} Hits the $12{,}000$-token limit without answering on $4/12$ runs.\par\vspace{1.8pt}
  \scriptsize \textbf{$B{=}512$ ($7/12$):} \suptileno{C}\suptileno{C}\suptileno{C}\suptileno{C}\,\suptileok{A}\suptileok{A}\suptileok{A}\suptileok{A}\,\suptileno{C}\suptileok{A}\suptileok{A}\suptileok{A}\par\vspace{1.8pt}
  \scriptsize \textbf{Prefix Credit Only ($7/12$):} \suptileok{A}\suptileno{B}\suptileok{A}\suptileok{A}\,\suptileno{C}\suptileok{A}\suptileno{C}\suptileok{A}\,\suptileno{C}\suptileno{C}\suptileok{A}\suptileok{A} \\
  \cmidrule(lr){1-1}\cmidrule(lr){2-2}\cmidrule(lr){3-3}
  \scriptsize \textcolor{qlbgreen!80!black}{\textbf{\method{} (Budget Cut, 2,063 tok):}} Reads the lower goldenrod curve cleanly: \emph{``top left shows \texttt{47 47} and to the right \texttt{47} alone\dots\ three points with value 47: $47{+}47{+}47 = 141$''} (\textbf{$5.3\times$ shorter}, \texttt{"141"}). &
  \scriptsize \textcolor{qlbgreen!80!black}{\textbf{\method{} ($12/12$ at $B{=}512$, $11/12$ Nat.):}} Compares band centers ($1.5 - 1.0 = 0.5\text{ MeV}$) and stops in \textbf{2,032 tok ($2.2\times$ shorter)}.\par\vspace{1.8pt}
  \scriptsize \textbf{$B{=}512$ ($12/12$):} \suptileok{D}\suptileok{D}\suptileok{D}\suptileok{D}\,\suptileok{D}\suptileok{D}\suptileok{D}\suptileok{D}\,\suptileok{D}\suptileok{D}\suptileok{D}\suptileok{D} &
  \scriptsize \textcolor{qlbgreen!80!black}{\textbf{\method{} ($12/12$ at $B{=}512$, $11/12$ Nat.):}} Identifies the empty band $[U_0{-}\Delta, U_0{+}\Delta]$ with $0/12$ cap hits in \textbf{2,413 tok ($3.0\times$ shorter)}.\par\vspace{1.8pt}
  \scriptsize \textbf{$B{=}512$ ($12/12$):} \suptileok{A}\suptileok{A}\suptileok{A}\suptileok{A}\,\suptileok{A}\suptileok{A}\suptileok{A}\suptileok{A}\,\suptileok{A}\suptileok{A}\suptileok{A}\suptileok{A} \\
  \bottomrule
\end{tabular}%
}

  \vspace{-2.0mm}
  \caption{\textbf{Additional qualitative success cases across external charts, main test plots, and unseen domains.}
    None of these cases appear in the main paper; tile strips show all $12$ student evaluations ($3$ training seeds $\times$ $4$ decoding seeds; green = correct, red = wrong).
    \emph{(a)} On an open-ended \textsc{ChartQA} question, Standard RFT overthinks for $10{,}848$ tokens and hallucinates a third \texttt{36} (\texttt{108}), whereas \method{} cleanly sums the three \texttt{47} points (\texttt{141}) in $2{,}063$ tokens ($5.3\times$ shorter).
    \emph{(b, c)} On nuclear-physics and unseen quantum-spectrum figures, \method{} achieves $12/12$ accuracy at $B{=}512$ and $11/12$ natural accuracy while cutting generation length by $2.2\times$ to $3.0\times$ and eliminating $12{,}000$-token cap hits.}
  \label{fig:supp-success}
  \vspace{2.0mm}
\end{figure*}

\begin{figure*}[p]
  \centering
  % Supplementary Failure Cases (3 real visual failure modes & boundaries)
\providecommand{\suptileok}[1]{\colorbox{qlbgreen!85!black}{\textcolor{white}{\texttt{\textbf{#1}}}}}
\providecommand{\suptileno}[1]{\colorbox{qlbverm!18}{\textcolor{qlbverm!90!black}{\texttt{#1}}}}
\resizebox{\textwidth}{!}{%
\setlength{\fboxsep}{1.1pt}%
\setlength{\tabcolsep}{4.5pt}%
\renewcommand{\arraystretch}{1.08}%
\begin{tabular}{@{}p{0.322\textwidth}p{0.322\textwidth}p{0.322\textwidth}@{}}
  \toprule
  \textbf{(a) Monochrome Line-Style Confusion} &
  \textbf{(b) Tight-Budget 10-Curve Cutoff} &
  \textbf{(c) Prefix-Horizon Symbolic Slip} \\
  \scriptsize \textsc{VeriSciQA} Main Test (Frozen Vision Encoder) &
  \scriptsize \textsc{VeriSciQA} Main Test ($B{=}512$ vs.\ Natural) &
  \scriptsize \textsc{MathVision} Testmini (Open-Ended Free-Form) \\
  \midrule
  \multicolumn{1}{c}{\includegraphics[height=2.45cm,keepaspectratio]{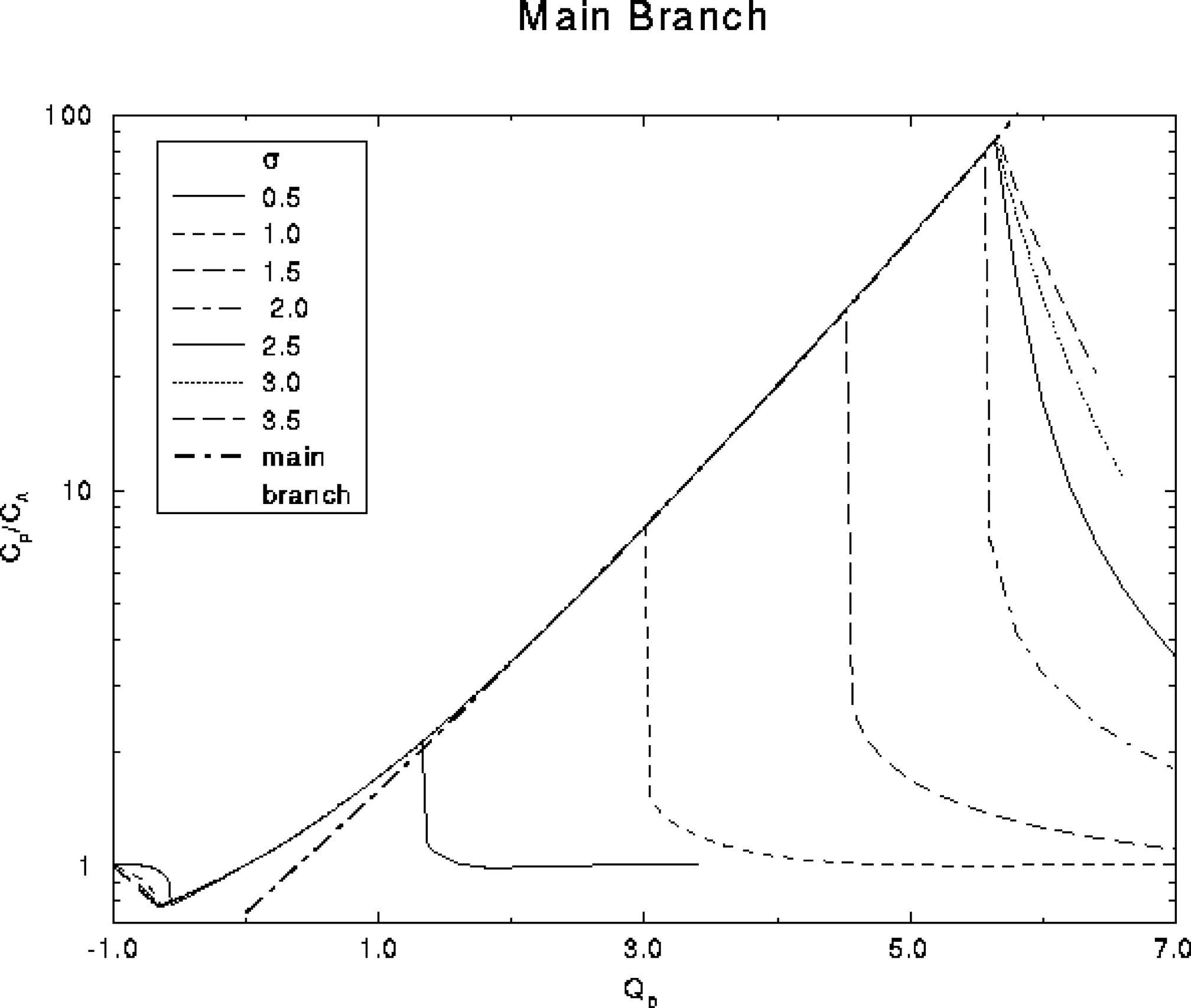}} &
  \multicolumn{1}{c}{\includegraphics[height=2.45cm,keepaspectratio]{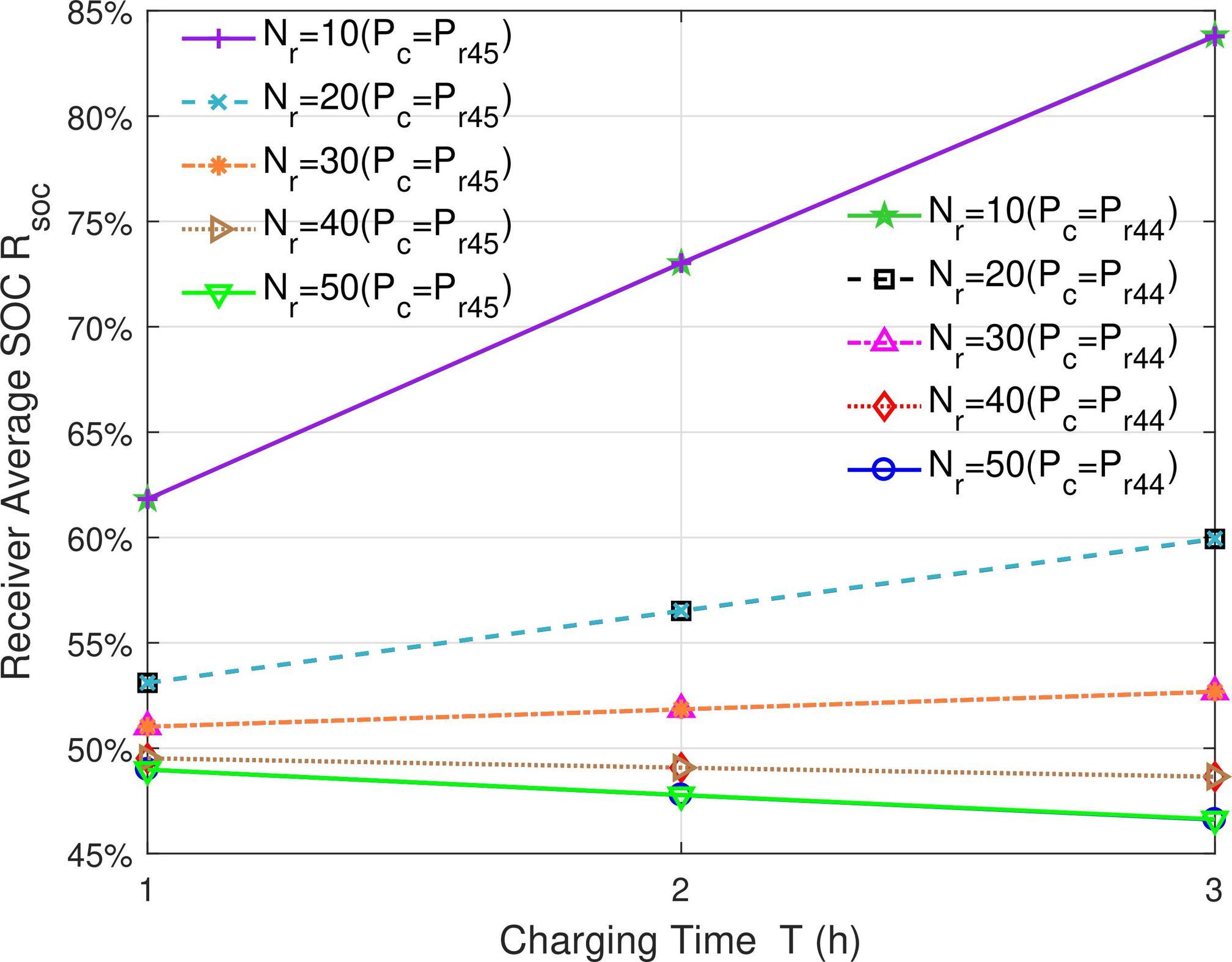}} &
  \multicolumn{1}{c}{\includegraphics[height=2.45cm,keepaspectratio]{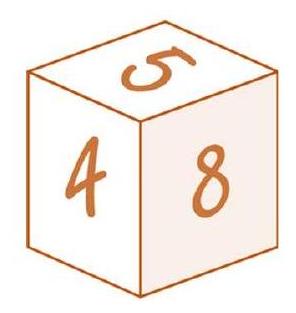}} \\
  \midrule
  \scriptsize \textbf{Q:} In the negative amplitude ($Q_0{<}0$) region, which Brill wave width $\sigma$ differs significantly? (\textbf{Gold: B}, $\sigma{=}0.5$) &
  \scriptsize \textbf{Q:} For which receiver numbers $N_r$ does average SOC decrease as charging time $T$ increases? (\textbf{Gold: B}, $40\text{ and }50$) &
  \scriptsize \textbf{Q:} Six distinct integers in $1\dots 9$ sit on a cube with equal opposite sums; which number can be opposite $8$? (\textbf{Gold: 3}) \\
  \midrule
  \scriptsize \textbf{Observed Behavior:} Standard RFT averages $6{,}310$ tokens ($1/12$ correct, $2/12$ cap hits). \method{} stops cleanly on all $12/12$ runs ($1{,}769$ mean tokens) and correctly zooms into $Q_0 \in [-1, 0]$, yet predicts \texttt{A} ($\sigma{=}3.0$) across runs. &
  \scriptsize \textbf{Forced $B{=}512$ ($1/12$):} \suptileno{D}\suptileno{D}\suptileno{D}\suptileno{D}\,\suptileno{D}\suptileno{D}\suptileno{D}\suptileok{B}\,\suptileno{D}\suptileno{D}\suptileno{D}\suptileno{D}\par\vspace{1.8pt}
  \scriptsize \textbf{Natural Finish ($10/12$):} \suptileok{B}\suptileok{B}\suptileok{B}\suptileok{B}\,\suptileno{D}\suptileok{B}\suptileok{B}\suptileno{D}\,\suptileok{B}\suptileok{B}\suptileok{B}\suptileok{B}\par\vspace{1.8pt}
  \scriptsize Checking all $10$ legend curves across two $P_c$ groups exceeds $512$ tokens, so forcing an answer at $512$ tokens cuts off elimination early. &
  \scriptsize \textbf{\method{} Trace ($3{,}072$ think tok, \texttt{finish=stop}):} Derives the right pairing in the prefix (\emph{``In this model: 8 is opposite 3, 4 is opposite 7, 5 is opposite 6''}), then closes \texttt{</think>} mid-check (\emph{``such that their opposites are \texttt{</think>}''}) and emits \texttt{"2"}. \\
  \cmidrule(lr){1-1}\cmidrule(lr){2-2}\cmidrule(lr){3-3}
  \scriptsize \textcolor{hvAmber}{\textbf{Takeaway:}} Prefix Credit eliminates tail looping, but resolving nearly identical monochrome dash styles requires vision-side adaptation (such as Vision LoRA in \cref{tab:ablation}a). &
  \scriptsize \textcolor{hvAmber}{\textbf{Takeaway:}} Unlike perceptual failure, this error comes from the external $512$-token test cap: under natural termination ($2{,}128$ mean tokens), \method{} solves $10/12$ runs. &
  \scriptsize \textcolor{hvAmber}{\textbf{Takeaway:}} Because \method{} is trained with a $3{,}072$-token prefix budget, multi-step symbolic puzzles that exceed $3{,}072$ tokens can trigger early stop-tag emission. \\
  \bottomrule
\end{tabular}%
}

  \vspace{-2.0mm}
  \caption{\textbf{Representative failure cases illustrating perceptual, tight-budget, and prefix-horizon boundaries.}
    \emph{(a)} Under a frozen vision encoder, \method{} stops cleanly ($1{,}769$ mean tokens, $12/12$ Finish) and locates the target $Q_0{<}0$ region, yet predicts \texttt{A} ($\sigma{=}3.0$) instead of \texttt{B} ($\sigma{=}0.5$) across overlapping monochrome dashes.
    \emph{(b)} On a dense $10$-curve single-panel plot, an external $512$-token cutoff interrupts curve-by-curve elimination ($1/12$ at $B{=}512$), whereas natural termination ($2{,}128$ tokens) recovers $10/12$ accuracy.
    \emph{(c)} On an open-ended \textsc{MathVision} cube puzzle, \method{} derives the right pairing (\emph{``8 is opposite 3''}) in its prefix, but naturally emits \texttt{</think>} at its learned $3{,}072$-token prefix horizon during final adjacency checking and outputs \texttt{2}.}
  \label{fig:supp-failure}
\end{figure*}

\paragraph{Multi-domain qualitative success analysis.}
\Cref{fig:supp-success} complements \cref{fig:case-studies} with three unseen cases: on \textsc{ChartQA} (\cref{fig:supp-success}a), Standard RFT spends $10{,}848$ tokens second-guessing repeated labels (\texttt{108}), whereas \method{} sums the three \texttt{47} points (\texttt{141}) in $2{,}063$ tokens ($5.3\times$ shorter); on Main and Unseen Reserve (\cref{fig:supp-success}b,c), all $12$ student runs confirm stronger early deduction ($12/12$ at $B{=}512$ vs.\ $5/12$--$8/12$) and zero $12{,}000$-token cap hits.

\paragraph{Diagnostic failure modes and boundaries.}
\Cref{fig:supp-failure} illustrates three scope boundaries from \cref{sec:conclusion}: (a) on overlapping monochrome dashes, Prefix Credit stops looping ($1{,}769$ vs.\ $6{,}310$ tok) but a frozen vision encoder still misbinds dash styles, motivating \texttt{Vision LoRA}; (b) on a $10$-curve plot, external $B{=}512$ cuts off elimination ($1/12$), whereas natural termination ($2{,}128$ tok) recovers $10/12$; and (c) on symbolic \textsc{MathVision} puzzles reaching the $3{,}072$-token budget, learned \texttt{</think>} emission can fire mid-check even after deriving the right pairing, motivating adaptive prefix budgets and multi-token open-ended readouts.

\end{document}